\documentclass[10pt,twocolumn]{article} 
\usepackage{simpleConference}
\usepackage{times}
\usepackage{graphicx}
\usepackage{amssymb}
\usepackage{url,hyperref}

\usepackage{bm}
\usepackage{amsmath}
\usepackage[caption=false,font=footnotesize]{subfig}
\usepackage{multirow}
\usepackage{adjustbox}
\usepackage{algorithm}
\usepackage[noend]{algpseudocode}
\usepackage{longtable}
\usepackage{supertabular,booktabs}
\usepackage{balance}
\usepackage{soul}
\usepackage{comment}
\usepackage[section]{placeins}

\algnewcommand\algorithmicforeach{\textbf{for each}}
\algdef{S}[FOR]{ForEach}[1]{\algorithmicforeach\ #1\ \algorithmicdo}
\algnewcommand{\algorithmicgoto}{\textbf{go to}}%
\algnewcommand{\Goto}[1]{\algorithmicgoto~\ref{#1}}%

\newcommand\Mark[1]{\textsuperscript#1}

\title{Generating an Overview Report over Many Documents}

\author{Jingwen Wang\Mark{1}, Hao Zhang\Mark{1}, Cheng Zhang\Mark{1}, Wenjing Yang\Mark{1}, Liqun Shao\Mark{2}, Jie Wang\Mark{3}\\
	\Mark{1},\Mark{3}University of Massachusetts Lowell, MA 01854\\
	\Mark{2}Microsoft Corporation,  Cambridge, MA 01854\\
	\Mark{1}\texttt{\{jingwen\_wang, hao\_zhang, cheng\_zhang, wenjing\_yang\}@student.uml.edu} \\
	\Mark{2}\texttt{liqun.shao@microsoft.com} \\
	\Mark{3}\texttt{wang@cs.uml.edu} \\
}

\begin{document}
	
\maketitle

\begin{abstract}
	How to efficiently generate an accurate, well-structured overview report (ORPT) over thousands of related documents is challenging. A well-structured ORPT consists of sections of multiple levels (e.g., sections and subsections). 
	None of the existing multi-document summarization (MDS) algorithms is directed toward this task.
	To overcome this obstacle, we present NDORGS (Numerous Documents' Overview Report Generation Scheme) that integrates text filtering, keyword scoring, single-document summarization (SDS),  topic modeling, MDS, and title generation to generate a coherent, well-structured ORPT.
	We then devise a multi-criteria evaluation method using techniques of text mining and multi-attribute decision making on a combination of human judgments, running time, information coverage, and topic diversity. We evaluate ORPTs generated by NDORGS on two large corpora of documents, where one is classified and the other unclassified. We show that, using Saaty's pairwise comparison 9-point scale and under TOPSIS, the ORPTs generated on SDS's with the length of 20\% of the original documents are the best overall on both datasets.
\end{abstract}

\section{Introduction}
\label{sec:introduction}

It is a challenging task to generate an accurate overview report  (ORPT) over a large corpus of related documents
in reasonable time. This task arises from the need of analyzing sentiments contained in a large number of documents.
By ``large'' we mean several thousand and by ''reasonable time'' we mean a few hours of CPU time on a commonplace computer. Moreover,
an ORPT should be of a reasonable size to save reading time. For a corpus of about 5,000 documents, for instance, an ORPT should not exceed 20 pages. 
Moreover, an OPT should be well-structured, organized in two or three levels for easier reading.
It should also include figures to highlight frequencies and trends of interesting entities such as names of people and organizations contained in the documents. 

Most of the early multi-document summarization (MDS) systems are designed to produce a summary for a handful of documents. 
For example, the MDS systems \cite{Christensen2013TowardsCM, Yasunaga2017GraphbasedNM,cao2017improving} are designed to handle the sizes of DUC datasets. The DUC-02, DUC-03, and DUC-04 datasets \cite{DUCdataset} each consists of only about five hundred documents, and provide bench-marking summaries for MDS tasks each consisting of 10 or fewer documents. 
Directly applying these MDS systems to generating an ORPT for a corpus containing thousands of documents
could possibly generate a proportionately longer and disorganized summary. More recently, T-CMDA \cite{Liu2018GeneratingWB} was devised to generate an English Wikipedia article on a specific topic over thousands of related documents. However, T-CMDA is still not suitable for generating a well-organized ORPT for a large corpus of documents containing multiple topics. 

To accomplish our task, we devise a document generation scheme called NDORGS, 
which stands for \textbf{N}umerous \textbf{D}ocuments' \textbf{O}verview \textbf{R}eport \textbf{G}eneration \textbf{S}cheme.
NDORGS is capable of generating a coherent and well-structured ORPT over a large corpus of input documents.
In particular, NDORGS first filters out noisy data, and then uses a suitable SDS algorithm 
to generate single-document summaries for each document with a length ratio $\lambda$ (e.g., $\lambda = 0.1, 0.2$, or 0.3) of the original document. For simplicity, we refer to a summary with a length ratio $\lambda$ of the original document as a $\lambda$-summary.
NDORGS then uses a suitable clustering algorithm 
to partition the corpus of
$\lambda$-summaries into two-level clusters. 
For each cluster of $\lambda$-summaries, NDORGS generates a section or a subsection using a suitable MDS algorithm, 
and uses a suitable 
title generation algorithm
to generate a title for that cluster.
Finally, NDORGS
generates an ORPT by integrating sections and corresponding subsections in 
the order of descending salience of clusters. 

We devise an evaluation method using human evaluations and text mining techniques to measure the quality of ORPTs generated by NDORGS, and use the running time to measure efficiency. 
We combine the following criteria in descending order of importance: human evaluations, running time, coverage, and diversity, and use multi-attribute decision-making techniques
to find the best value of $\lambda$. 

Human evaluations follow the standard criteria given in DUC-04 \cite{DUCdataset},
while
coverage and diversity are evaluated by comparing, respectively, keyword scores  in 
an ORPT and the original corpus of documents, and the symmetric difference of clusters of keywords. We use Saaty's pairwise comparison 9-point scale and TOPSIS to determine the best overview. We apply NDORGS to 
a corpus of 2,200 classified BBC news articles, thereafter referred to as BBC News; 
and a corpus of 5,300 unclassified articles extracted from Factiva \cite{Factiva} under the keyword search of ``Marxism" from the year of 2008 to the year of 2017, hereafter referred to as Factiva-Marx, for a project of analyzing public sentiments. We show that for both datasets, using Semantic WordRank \cite{2018SemanticWordRank} to generate single-document summaries, Latent Dirichlet Allocation (LDA) \cite{Blei2003LatentDA} to cluster summaries, 
GFLOW \cite{Christensen2013TowardsCM} to generate a summary over a cluster of summaries, and DTATG \cite{DTATGShao2016} to generate a title for a section (or subsection),
the ORPTs generated on 0.2-summaries provide the best overall ORPTs. Sensitivity analysis shows that this result is stable.

The rest of this paper is organized as follows: In
Section \ref{sec:related} we describe related work,
and in
Section \ref{sec:design} we present the architecture of NDORGS and the algorithms.
We present experiment and evaluation results in  
Section \ref{sec:experiments}, and trending methods and sample figures in Section \ref{sec:trending}. We conclude the paper
in Section \ref{sec:conclusion}.

\section{Related Work}
\label{sec:related}

We discuss related work on topic modeling, text summarization, and title generation.

\subsection{Topic modeling}

Topic modeling 
partitions documents into topic clusters with each cluster representing a unique topic. 
Latent Dirichlet Allocation (LDA) \cite{Blei2003LatentDA} and Spectral Clustering (SC) \cite{ng2002spectral}
are popular topic clustering methods. 
LDA treats each document in a corpus as a mixture of latent topics that generate words for each (hidden) topic with certain probabilities. As a probabilistic algorithm, LDA may produce somewhat different clusters on the same corpus of documents on different runs. It also
needs to predetermine the number of topics. 
SC is a deterministic and faster clustering algorithm, which also
needs to preset the number of topics.
It uses eigenvalues of an affinity matrix to reduce dimensions. It then uses k-means to generate clusters over eigenvectors
corresponding to the $k$ smallest eigenvalues.

Other clustering methods, such as 
PW-LDA \cite{li2018lda}, Dirichlet multinomial mixture \cite{yin2014dirichlet}, and neural network models \cite{xu2015short}, are targeted at corpora of short documents such as
abstracts of scientific papers, which are not suited for our task. 
The documents we are dealing with are much longer. Even if we use summaries to represent the original documents, a summary may still be significantly longer than a typical abstract. 
We note that the k-NN graph model \cite{lulli2015scalable} may also be used
for topic clustering. 

Methods for keyword/phrase extraction include k-Means \cite{hartigan1979algorithm}, TF-IDF \cite{Salton1988TermWeightingAI}, TextRank \cite{Mihalcea2004TextRank}, RAKE \cite{Rose2010AutomaticKE}, and AutoPhrase \cite{shang2018automated}.

\subsection{Text summarization}

Text summarization includes SDS, MDS, hierarchical summarization, and structural summarization.

\subsubsection{Single-document summarization}

SDS algorithms have been studied intensively and extensively for several decades 
(for example, see \cite{Mihalcea2004TextRank,yogatama2015extractive,cao2016attsum,nallapati2017summarunner,2018SemanticWordRank}).
As it is today, unsupervised, extractive algorithms are still more efficient, more accurate, and more flexible 
than supervised or abstractive summarization. 

Among unsupervised extractive algorithms,
the Semantic WordRank (SWR) \cite{2018SemanticWordRank} algorithm is currently the best in terms of both accuracy and efficiency.
Built on a  
weighted word graph with semantic 
and co-occurrence edges, 
SWR scores sentences using an article-structure-biased PageRank algorithm \cite{page1999pagerank} 
with a Softplus function elevation adjustment, and promotes topic diversity using spectral subtopic clustering under the Word-Movers-Distance metric. SWR
outperforms all previous algorithms (supervised and unsupervised) over DUC-02 under the standard ROUGE-1, ROUGE-2, and ROUGE-SU4 measures. 
Over the SummBank dataset, SWR 
outperforms each of the three human annotators and compares favorably with the combined performance of the three annotators.

\subsubsection{Multi-document summarization.}

An MDS algorithm takes several documents as input 
and generates a summary as output. 
Most MDS algorithms are algorithms of selecting sentences. 
Sentences may be ranked using features of term frequencies, sentence positions, and keyword co-occurrences \cite{hong2014repository, Mihalcea2004TextRank}, among a few other things. 
Sentences may be selected using algorithms such as graph-based lexical centrality LexRank \cite{erkan2004lexrank}, centroid-based clustering \cite{Radev2004CentroidbasedSO},  
Support Vector Regression \cite{Li2007MultidocumentSU},
syntactic linkages between text \cite{wang2016exploring}, 
and Integer Linear Programming \cite{gillick2009scalable, li2013using}. 
Selected sentences may be reordered to improve coherence using 
probabilistic methods \cite{Lapata2003ProbabilisticTS,Nayeem2017EwO}. 

Among all the MDS algorithms, the GLFOW algorithm \cite{Christensen2013TowardsCM} 
is focused on sentence coherency. 
GFLOW is an unsupervised graph-based method that extracts a coherent summary from multiple documents by selecting and reordering sentences to balance coherence and salience over an approximate discourse graph (ADG) of sentences.
An ADG considers sentence discourse relations among documents based on relations such as deverbal noun reference, event/entity continuation, discourse markers, sentence inference, and co-reference mentions. 
In an ADG graph, each node represents a sentence. Two nodes are connected if they have one of the aforementioned sentence relations. Edge weight is calculated based on the number of sentence relations between two sentences. 

Based on GFLOW, Yasunaga et al. \cite{Yasunaga2017GraphbasedNM} devised a supervised neural network model that combines Personalized Discourse Graph (PDG), Gated Recurrent Units (GRU), and Graph Convolutional Network (GCN) \cite{kipf2016semi} to 
rank and select sentences . 
TCSum \cite{cao2017improving} is another neural network model that leverages text classification to improve the quality of multi-document summaries. 
However, neural network methods require large-scale training data to obtain a good result. 

More recently Liu et al. \cite{Liu2018GeneratingWB} devised a large-scale summarization method named T-DMCA to generate an English Wikipedia article. T-DMCA combines  extractive summarizations and abstractive summarizations trained on a large-scale Wikipedia dataset to summarize the text. While T-CMDA is capable of creating summaries with specified topics as Wikipedia article, it can hardly generate an overview report for a large corpus of documents containing multiple topics. Moreover, T-DMCA fails on ranking its sub-topics based on their salience.

\subsubsection{Hierarchical and structural summarization.}

Hierarchical summarization and structural summarization are two approaches to 
enhancing SDS and MDS algorithms.
Buyukkokten et al.
\cite{Buyukkokten:2001:SWP} and Otterbacher et al. \cite{Otterbacher06newsto} 
devised algorithms for generating a hierarchical summary of a single document. 
A hierarchical summarizer for Twitter tweets based on Twitter-LDA summarizes the news tweets into a flexible, topic-oriented hierarchy \cite{gao2012joint}.
SUMMA \cite{christensen2014hierarchical} is a system that creates coherent summaries hierarchically in the order of time, locations, or events.
These methods focus on single documents or short texts, or  
require documents be written with a certain predefined structure template, making them unsuited for our task.

Structured summarization algorithms first
identify topics of the input documents. 
Sauper et al. \cite{Sauper:2009:AGW} presented an overview generation system that uses a high-level structure of human-authored documents to generate a topic-structure 
multi-paragraph overview with domain-specific templates. 
Li et al. \cite{Li:2010:GTE} developed
a summary template generation system \cite{Li:2010:GTE} based on an entity-aspect LDA model to cluster sentences and words and generate sentence patterns to represent topics. Autopedia \cite{Yao:2011:AAD} is a Wikipedia article generation framework that selects Wikipedia templates as article structures. 

\subsection{Title generation}

Generating an appropriate title for a block of text is an important task for generating 
an overview report of multilevel structure, as sections and subsections each needs a title.
Alexander et al. \cite{Alexander:2015:NAMASS} attempted to generate an abstractive sentence summary as title for a given document. Ramesh et al. \cite{Ramesh:2016:S2STS} devised
an attentional encoder-decoder neural-network model used for machine translation \cite{Bahdanau:2014:NMTJL} to generate a title for short input sequences (one or two sentences). 
Shao and Wang  \cite{DTATGShao2016} devised DTATG for automatically generating an adequate title for a given block of text using dependency tree pruning. In particular, DTATG first extracts a few critical sentences that convey the main meanings of the text
and are more suitable to be converted into a title. Each critical sentence is given a numerical ranking score. It then constructs a dependency tree for each of these sentences, and 
trims certain non-critical branches from the tree to form a new sentence that
retains the original meanings and is still grammatically 
sound.
DTATG uses a title test consisting of a number of rules to determine if a sentence is suited for a title.
If a trimmed sentence passes the title test, then it becomes a title candidate.
DTATG selects the title candidate with the highest ranking score as the final title.
Evaluations by human judges showed that DTATG can generate adequate titles.

\section{Description of NDORGS}
\label{sec:design}

\begin{figure}[h]
	\centering 
	\includegraphics[width=\columnwidth]{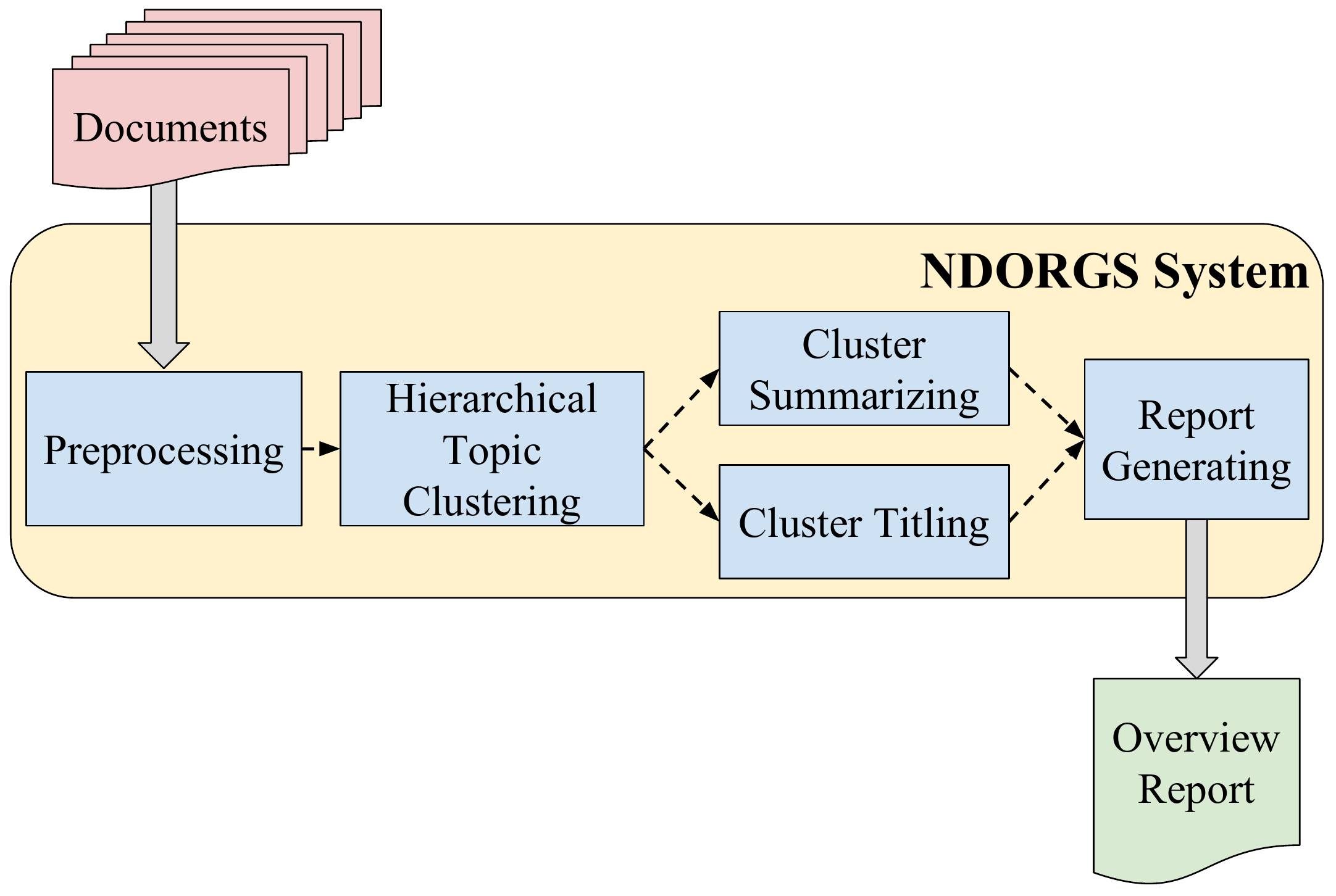}
	\caption{NDORGS architecture}
	\label{fig:ndorg}
\end{figure}

NDORGS (see Fig. \ref{fig:ndorg}) consists of five modules: (1) Preprocessing. (2) Hierarchical Topic Clustering.
(3) Cluster Summarizing. (4) Cluster Titling. (5) Report Generating.

\paragraph{\textbf{Step 1. Preprocessing (PP)}}

The PP module performs text filtering and SDS.
In particular, it first determines what language an input document is written in, eliminates irrelevant text (such as non-English articles, URLs, and duplicates),
and extracts, for each document, the title and subtitles (if any), publication time, publication source, and the main content. 
It also removes any ``interview'' type of articles because contents in interviews are too subjective and the structure of an interview is essentially a different genre.
It then generates a summary of appropriate length for each document. 
Extracting summaries is necessary for speeding up the process and 
is sufficient for generating a good overview
because only a small part of the most important content of an article 
will ultimately contribute to the overview report (also noted in \cite{Liu2018GeneratingWB}).

\paragraph{\textbf{Step 2. Hierarchical Topic Clustering (HTC)}}
\label{sec:HTC}

\begin{figure}[h]
	\centering
	\includegraphics[width=\columnwidth]{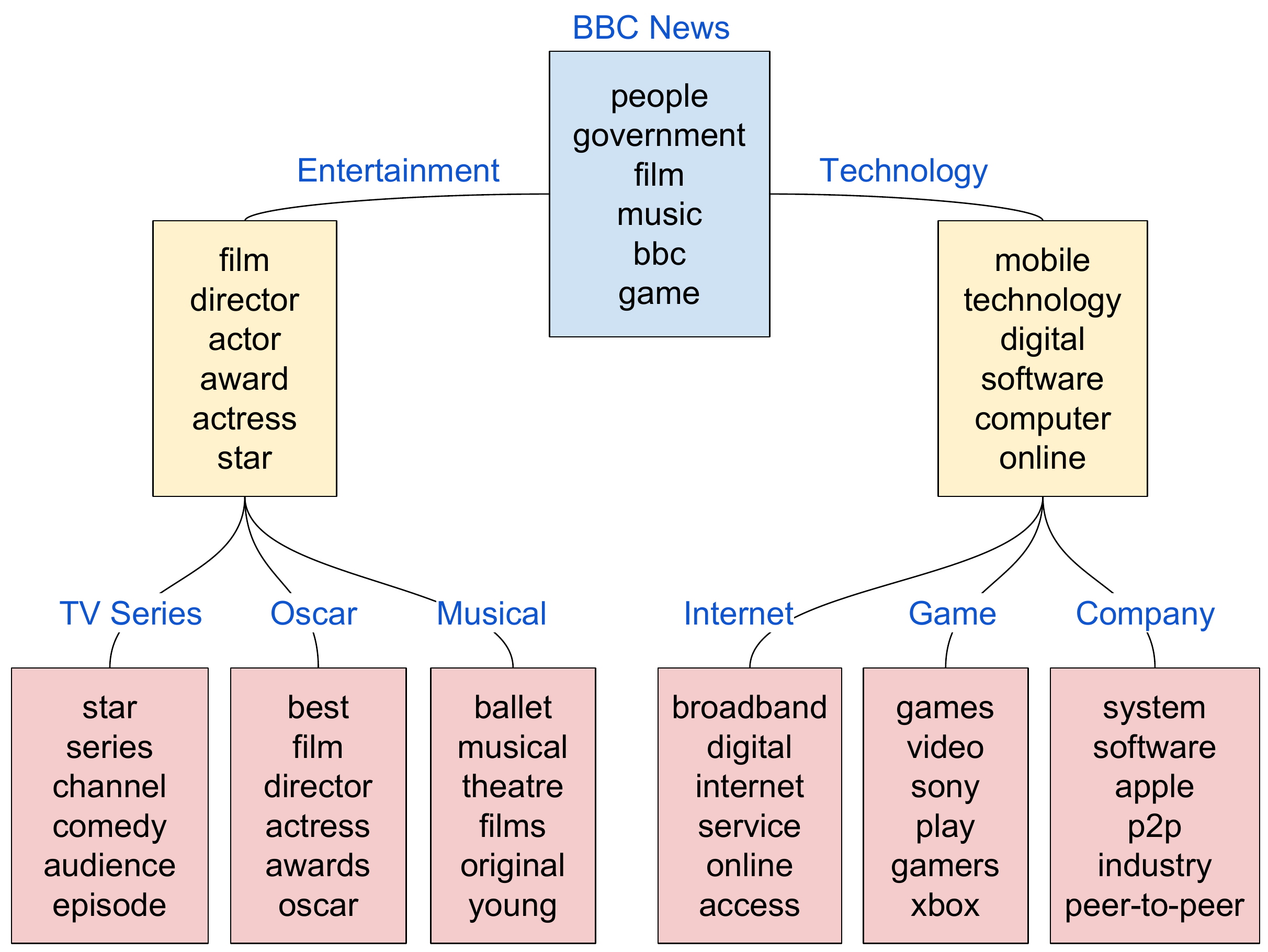}
	\caption{A subtree of topic clusters learned from BBC News dataset}
	\label{fig:bbc-topics}
\end{figure}

Multiple topics are expected over a large corpus of documents, where each topic may further contain subtopics.
Fig. \ref{fig:bbc-topics} is an example of a hierarchical topic subtree over BBC News. In this subtree, each node is a topic with six most frequent words under the topic. The root illustrates the most frequent words of the corpus. 
The first level topic cluster contains two subtopics: 
\textit{Entertainment} and \textit{Technology}. When words such as ``series'', ``comedy'', and ``episode'' under topic \textit{Entertainment} are discovered for a substantial number of times, a  subtopic of \textit{TV Series} may be detected. The hierarchically structured topic clusters provide detail information about topic relationships contained in a large corpus of documents.   

HTC is designed to capture topic relations. 
In particular, the HTC module partitions a corpus of
documents into multilevel topic clusters and ranks the topic clusters based on their salience. We present a two-level topic clustering algorithm in this paper (three or more level topic clustering is similarly designed). 
The input can be either the original documents or the summaries generated in Step 1.
Note that clustering on the original documents and clustering on the summaries of the
documents may result in different partitions (see Section \ref{label:para_setting_lambda}).  
For simplicity, we use ``documents" to denote both. 

In particular, HTC first partitions documents into $K$ clusters, denoted by 
${\mathcal C} = \{C_1, C_2, \ldots, C_K\}$, using a clustering method such as LDA or SC. 
These are referred to as the top-level clusters. 
For each top-level cluster $C_i$, 
if $|C_i| > N$, where $|C_i|$ denotes 
the number of documents contained in $C_i$ and $N$ is a preset number (for example, $N = 100$), then HTC 
further partitions $C_i$ into $K_i$ sub-clusters 
as the second-level clusters, where $K_i = 1+\lfloor |C_i|/N\rfloor$.
If in this new clustering, all but one cluster are empty, then this means that
documents in $C_i$ cannot be further divided into sub-topics. In this case,
we sort the documents in $C_i$ in descending order of document scores and 
split $C_i$ evenly into $K_i$ clusters (except the last one).

For a second-level cluster $C_{ij}$ of $C_i$, if $|C_{ij}| > N$, we may further create a third-level sub-clustering by clustering $C_{ij}$ or simply splits $C_{ij}$ evenly
into $K_{ij} = 1+ \lfloor |C_{ij}|/N\rfloor$ clusters (except the last one), still at the second-level.
Note that at each level of clustering, it is possible to have
empty clusters. 

Assume that cluster $C$ consists of $n$ documents, denoted by $d_i$, where
$i =1,2,\ldots,n$.
Let
$p_i$ denote the probability that $d_i$ belongs to 
cluster $C$. (Such a probability can be easily computed using LDA.)
We define the score of cluster $C$ using the following empirical formula:
\[ S_{C} = \frac{1}{n^2}\sum_{j=i}^{n} 2^{p_i}. \]

\begin{figure*}[t!]
	\centering
	\includegraphics[width=\linewidth]{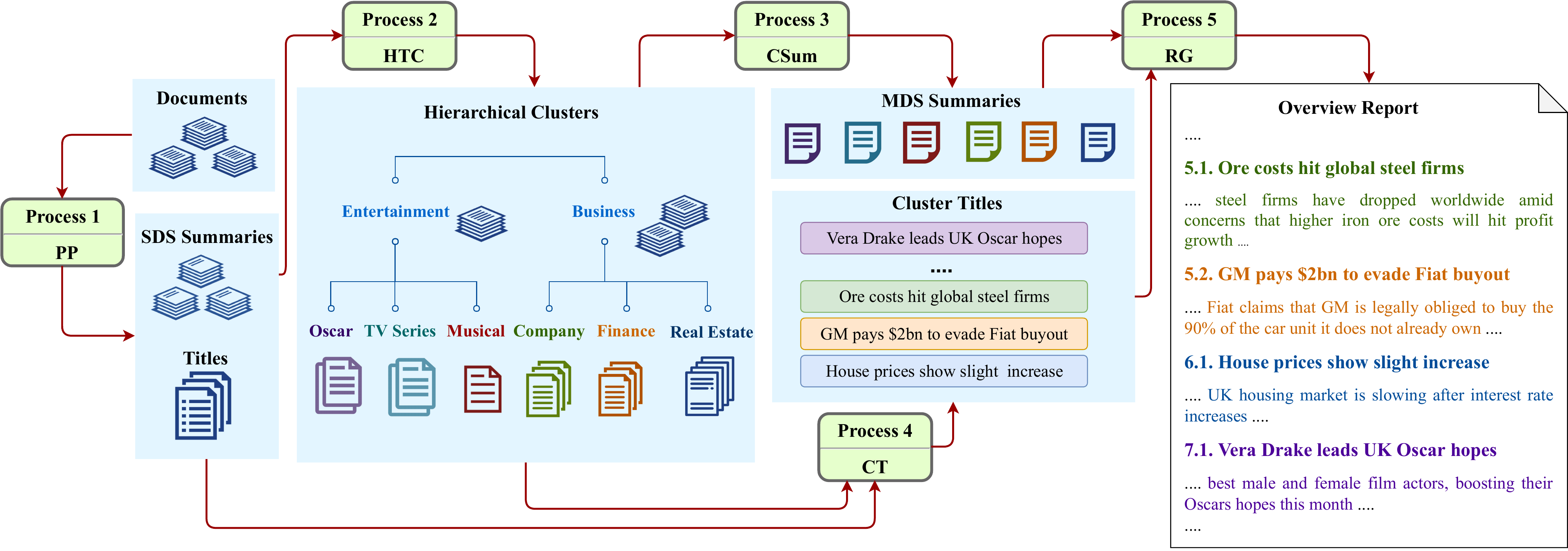}
	\caption{An example of NDORGS processing}
	\label{fig:NDORG_eg}
\end{figure*}

\paragraph{\textbf{Step 3. Cluster Summarizing (CSum)}}
\label{sec:CSum}
The CSum module generates a coherent summary of an appropriate length for all SDS summaries in a given cluster using an MDS algorithm. This cluster is typically at the second level except when a first-level cluster does not have a second-level cluster (in this case, the cluster is at the first level).
In particular, for each second-level cluster, CSum takes its corresponding SDS summaries as input and uses an appropriate MDS to produce a summary of a suitable length.

\paragraph{\textbf{Step 4. Cluster Titling (CT)}}
\label{sec:CT}

A series of good section headings is one of the most important components in a clear and well-organized overview reports. Since an overview report contains multiple topics, section headings help readers identify the main point of each section in the overview report.

The CT module creates appropriate titles for both first-level and second-level clusters
using a title generation algorithm.

\paragraph{\textbf{Step 5. Report Generating (RG)}}
As mentioned in Step 2, 
each cluster 
has a topic score 
and each document summary 
has a corresponding probability score. 
A higher cluster score represents a more significant topic. 
The RG module reorders clusters at the same level according to cluster scores 
in descending order and generates an overview report of up to $K$ top-level sections,
where each top-level section may also contain up to $K_i$ second-level sections.
Each section and subsection has a section title, and each subsection contains
an MDS summary. 

\subsubsection*{Example}
Fig. \ref{fig:NDORG_eg} is an example of the processes of NDORGS with an input of 15 documents.
The PP module determines what language an input document is written in, 
eliminates irrelevant text (such as non-English articles, URLs, and duplicates), 
and extracts information including the title and the main content. It then generates an SDS summary for each main content. The HTC module groups the fifteen summaries into two-level hierarchical topic clusters, with two top-level clusters, where the cluster \textit{Entertainment} consists of five document summaries and the cluster \textit{Business} consists of ten summaries. 
Both of the top-level clusters have three second-level clusters, where each second-level cluster consists of a number of SDS summaries and is labeled with a unique color. 
In Step 3, CSum extracts a coherent MDS summary for 
each second-level cluster summaries and the first-level cluster if there are no second-level
clusters. 
In Step 4, NDORGS produces a section title for each cluster and sub-cluster. 
Finally, the RG module rearranges the summaries and section titles based on their salience to generate an overview report with a multilevel structure.


\section{Evaluations}
\label{sec:experiments}

We describe datasets, selections of algorithms, and parameter setting for evaluating NDORGS.

\subsection{Datasets}
\label{sec:datasets}
We use the following two types of corpora: 

\begin{enumerate}
	\item The \textbf{BBC News} dataset \cite{greene06icml}  of 2,225 classified articles stemmed from BBC News in the years of 2004 and 2005 labeled in business, entertainment, politics, sports, and technology.
	\item The \textbf{Factiva-Marx} dataset\footnote{The Factiva-Marx dataset is available at http://www.ndorg.net.} \cite{MarxDataset} of 5,300 unclassified articles extracted from Factiva under the keyword search of ``Marxism''. 
\end{enumerate}
The statistics of these two datasets are shown in Table \ref{table:datasets}.

\begin{table*}[t]
	\centering
	\caption{Size comparisons between different datasets}
	\label{table:datasets}
	\resizebox{6.0in}{!}{%
		\begin{tabular}{ c c c c c c  }
			\hline
			\textbf{Dataset} & \textbf{\# of docs}  & \textbf{avg. \# of docs / task} & \textbf{\# of tokens} & \textbf{avg. \# of tokens / doc} & \textbf{vocabulary size} \\
			\hline
			Factiva-Marx     & $5,300$ & $5,300$ & $1.09 \times 10^7$ & $2,100$ & $3.89 \times 10^5$ \\
			BBC News & $2,225$ & $2,200$ & $8.5 \times 10^5$  & $380$ & $6.56 \times 10^4$ \\
			\hline
		\end{tabular} %
	}
\end{table*}

\subsection{Selection of algorithms}

We use the state-of-the-art Semantic WordRank \cite{2018SemanticWordRank} as the SDS algorithm and GFLOW \cite{Christensen2013TowardsCM} as the MDS algorithm for NDORGS.
In particular, GFLOW needs to solve the following ILP problem:
\begin{eqnarray*}
	\mbox{maximize} & & 
	Sal(X) + \alpha Coh(X) - \beta |X| \\
	\mbox{subject to} && \sum_{x_i \in X} l(x_i) < B, \\
	&&\forall x_i, x_j \in X: rdt(x_i, x_j) = 0,
\end{eqnarray*}
\noindent
where variable $X$ is a summary, $|X|$ is the number of sentences in the given summary, $Sal(X)$ is the salience score of $X$, $Coh(X)$ is the coherence score of $X$,
$rdt(x_i,x_j)$ is a  measure between two sentences $x_i$ and $x_j$, and
$l(x_i)$ is the length of sentence $x_i$. Parameters $\alpha$, $\beta$, and $\lambda$ are learned using the DUC-03 dataset.

We investigate LDA \cite{Blei2003LatentDA} and SC \cite{ng2002spectral} and determine
that LDA is more appropriate. We then use DTATG \cite{DTATGShao2016} 
to generate a title for each section and subsection.

\subsection{Parameter settings}
\label{sec:paraSetting}
We determine empirically the setting of parameters that lead to the best 
overall performance for both BBC News and Factiva-Marx. 
For each dataset, NDORGS produces three ORPTs\footnote{The six ORPTs generated by NDORGS are available at http://www.ndorg.net.}
corresponding to three $\lambda-$summaries, where $\lambda \in \{0.1, 0.2, 0.3\}$. 
The parameter settings are listed below: 
\begin{enumerate}
	\item  In the PP module, NDORGS generates single-document summaries for each document with the length ratio $\lambda = 0.1, 0.2$, and 0.3. 
	\item  In the HTC module, NDORGS creates two-level topic clusters using LDA. To achieve a higher topic clustering accuracy, we set the number $K$ of the top-level clusters to $K=9$ as suggested in Section \ref{sec:compare_f1} (see Fig. \ref{fig:bbc-baseline-f1}). To generate the second-level clusters, we set $N=200$ to determine if a sub-cluster should be further divided (recall that
	if a cluster contains more than $N$ documents, a further division will be performed).
	The number of second-level clusters is automatically determined by NDORGS using the method mentioned in Step 2 from 
	Section \ref{sec:HTC}. 
	\item  In the CSum module, NDORGS uses GFLOW to produce cluster summaries. The length $l$ of an MDS summary of (nonempty) cluster $C$ is determined by 
	\begin{align*}
	l = \left\{ \begin{array}{ll}
	150 \cdot \lfloor |C|/10\rfloor + 300,  & \mbox{if }  |C|<70, \\
	200 \cdot \lfloor |C|/10\rfloor, & \mbox{if }  |C|\geq 70.
	\end{array}\right. 
	\end{align*}
	\item  In the CT module, NDORGS applies DTATG to generate a title for each cluster and sub-cluster. 
	\item  In the RG module, NDORGS reorders clusters at the same level according to cluster scores $S_{C}$ defined in Section \ref{sec:HTC}. For each level of clusters, if a cluster contains less than 70 documents, then we consider this cluster a minor topic. 
	NDORGS merges such cluster MDS's into a section under the title of ``Other Topics'',
	where the MDS's are each listed as a bullet item, in descending order of cluster scores. 
\end{enumerate}

To achieve the best performance of NDORGS, it is critical to determine the SDS length ratio $\lambda$ and the top-level number $K$ of clusters. In
Section \ref{label:para_setting_k} and Section \ref{label:para_setting_lambda} we describe  experiments for deciding these parameters.

\subsection{Text clustering evaluations for deciding $K$}
\label{label:para_setting_k}

Let $D$ be a corpus of text documents. Suppose that we have a gold-standard partition of $D$ 
into $K$ clusters $C=\{C_1, C_2,\ldots,C_K\}$, and a clustering algorithm generates $K$ clusters, denoted by $A=\{A_1, A_2,\ldots,A_K\}$. We rearrange these clusters so that the symmetric difference of $C_i$ and $A_i$, denoted by $\Delta(C_i,A_i)$, is minimum,
where $\Delta(X,Y)=|X \cup Y| - |X \cap Y|$. That is, for all
$1 \leq i \leq K$, 
$\Delta(C_i,A_i) = \min_{1 \le j \le k}\Delta(C_i,A_j).$
We define CSD F1-score for $A$ and $C$ as follows, where CSD stands for \textbf{C}lusters \textbf{S}ymmetric \textbf{D}ifference: 
\begin{align*}
F_1(A, C) &= \frac{1}{K}\sum_{i=1}^{K}{F_1(A_i, C_i)}, \\
F_1(A_i, C_i) &= \frac{2P(A_i,C_i)R(A_i,C_i)}{P(A_i,C_i)+R(A_i,C_i)}, 
\end{align*}
with $P$ and $R$ being precision and recall defined by
$P(A_i, C_i) = \frac{|A_i \cap C_i|}{|A_i|}$ and
$R(A_i, C_i) = \frac{|A_i \cap C_i|}{|C_i|}$.
Clearly, $F1(A, C) \le 1$ and $F1(A, C) = 1$ is the best possible.

Let HLDA-D, HSC-D, HLDA-S, and HSC-S denote, respectively, 
the algorithms of applying HLDA and HSC on original documents and 0.3-summaries generated by Semantic WordRank \cite{2018SemanticWordRank}.

\begin{figure}[h!]
	\centering
	\includegraphics[width=\columnwidth]{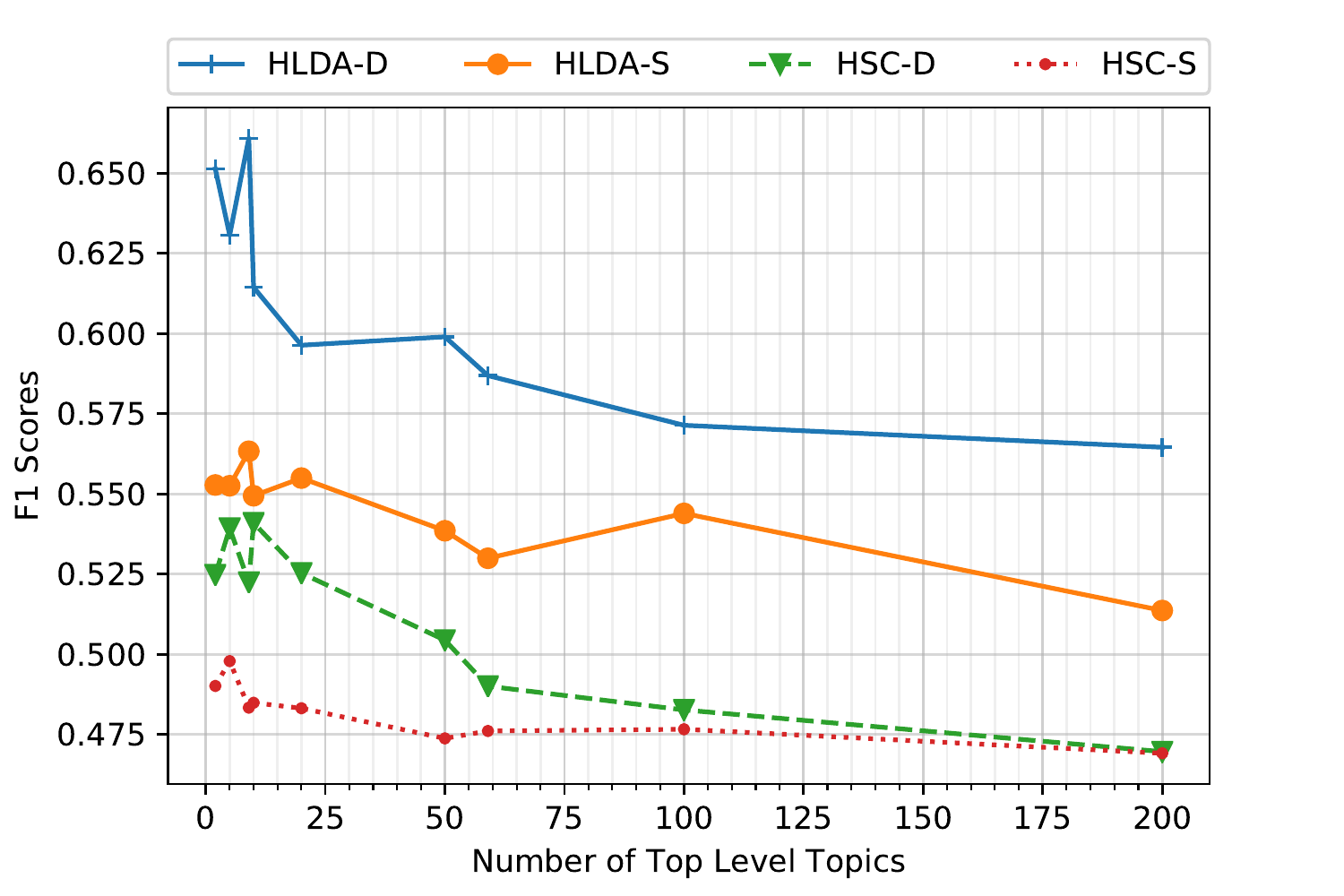}
	\caption{CSD F1-scores on the labeled BBC News corpus} 
	\label{fig:bbc-baseline-f1}
\end{figure}

\paragraph{\textbf{Comparisons of clustering quality}}\label{sec:compare_f1}
Fig. \ref{fig:bbc-baseline-f1} compares the CSD F1-scores of HLDA-D, HLDA-S, HSC-D, and
HSC-S over the labeled corpus of BBC News articles.
We can see that HLDA-D is better than HLDA-S, which is better than HSC-D, and HSC-D is better than HSC-S. All of these algorithms have the highest CSD F1-scores when the number of top-level topics $K=9$.  This is in line with a general experience that the number of top-level sections in an overview report should not exceed 10.

\begin{figure}[h!]
	\centering
	\subfloat[CSD F1-scores on BBC News]{
		\includegraphics[width=\columnwidth]{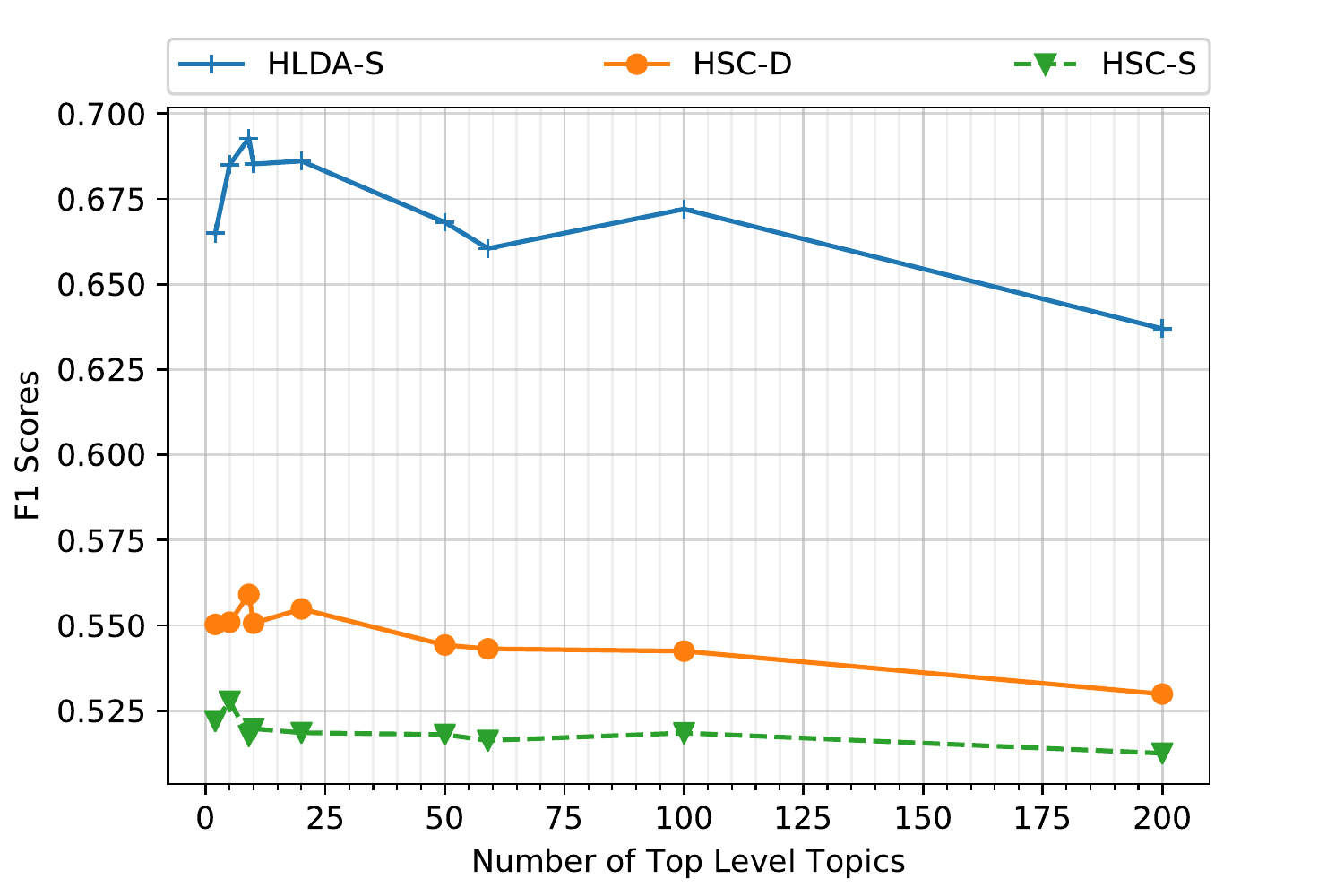}
		\label{fig:bbc_HLDAbaseline_f1}
	}
	\hfill
	\subfloat[CSD F1-scores on Factiva-Marx]{
		\includegraphics[width=\columnwidth]{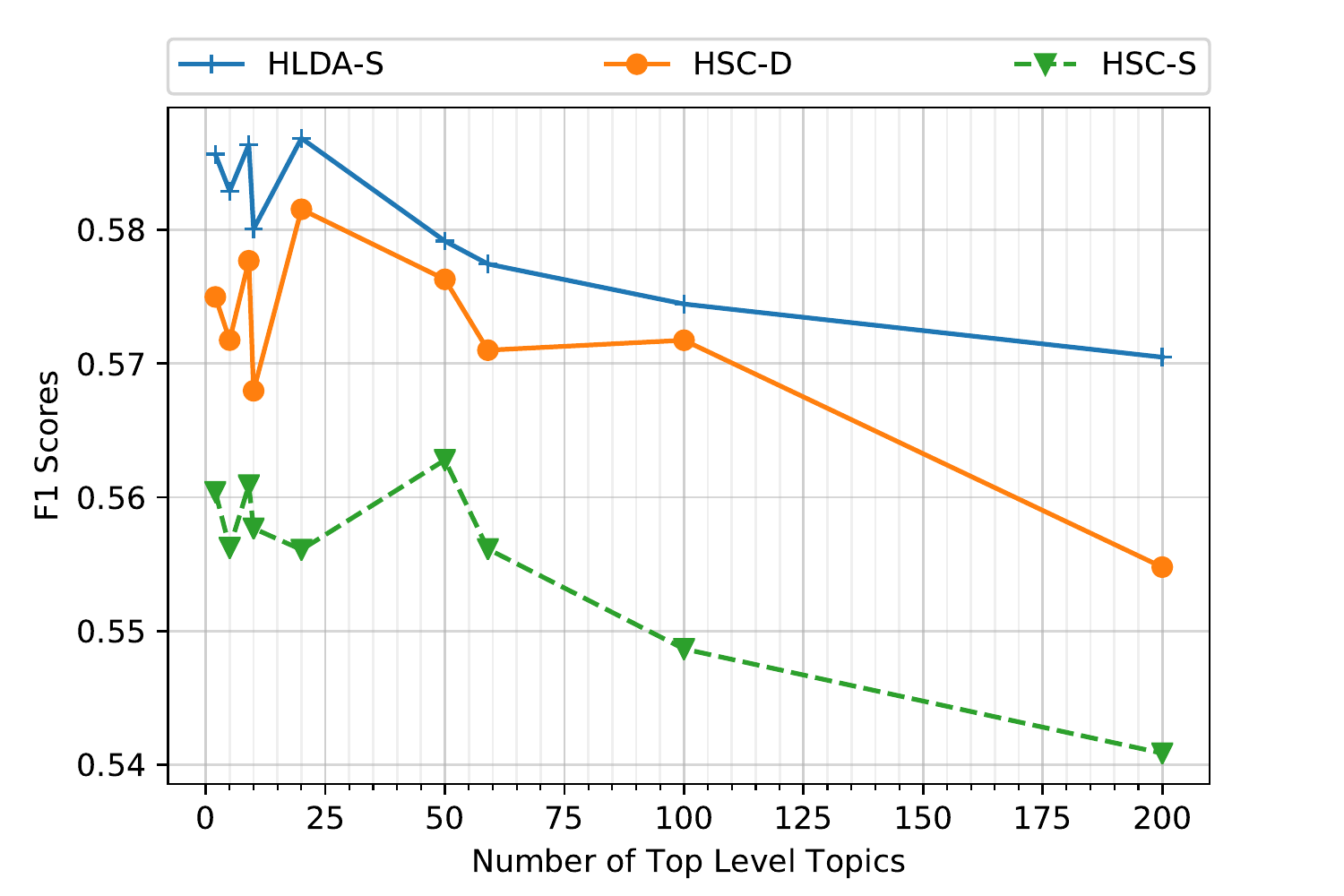}
		\label{fig:marx_HLDAbaseline_f1}
	}
	\caption{\small CSD F1-scores against HLDA-D}
	\label{fig:hlda_hsc_hlda_baseline_f1}
\end{figure}

We observe that HLDA-D offers the best accuracy. Thus, 
we will use the clustering generated by HLDA-D as the baseline for comparing CSD F1 scores.
Fig. \ref{fig:bbc_HLDAbaseline_f1} and Fig. \ref{fig:marx_HLDAbaseline_f1}
depict the comparison results of HLDA-S, HSC-D, and HSC-S against HLDA-D on
BBC News and Factiva-Marx.

\begin{figure}[h]
	\centering
	\subfloat[Running time of clustering the BBC News corpus]{
		\includegraphics[width=\columnwidth]{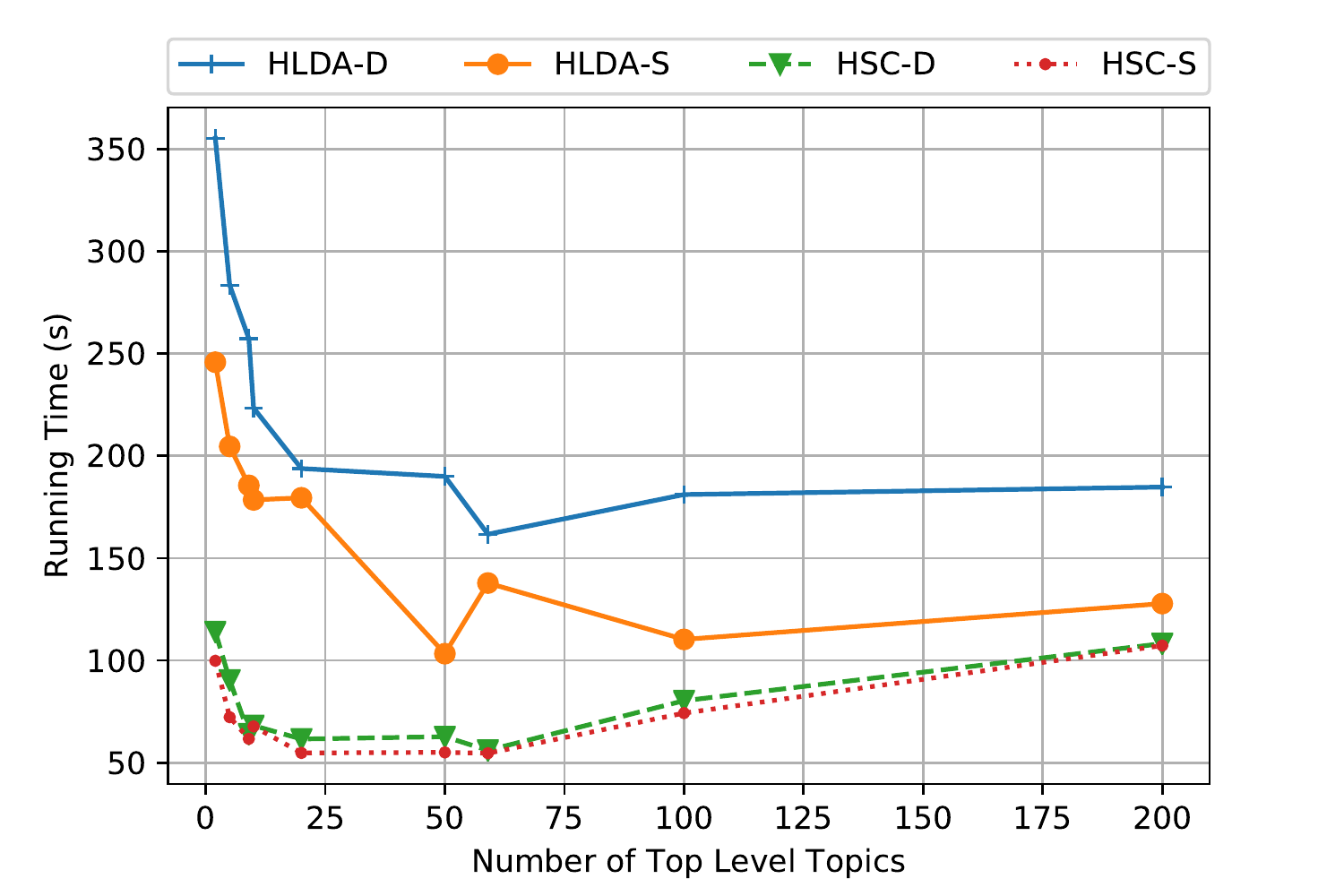}
		\label{fig:bbc_time}
	}
	\hfill
	\subfloat[Running time of clustering the Factiva-Marx corpus]{
		\includegraphics[width=\columnwidth]{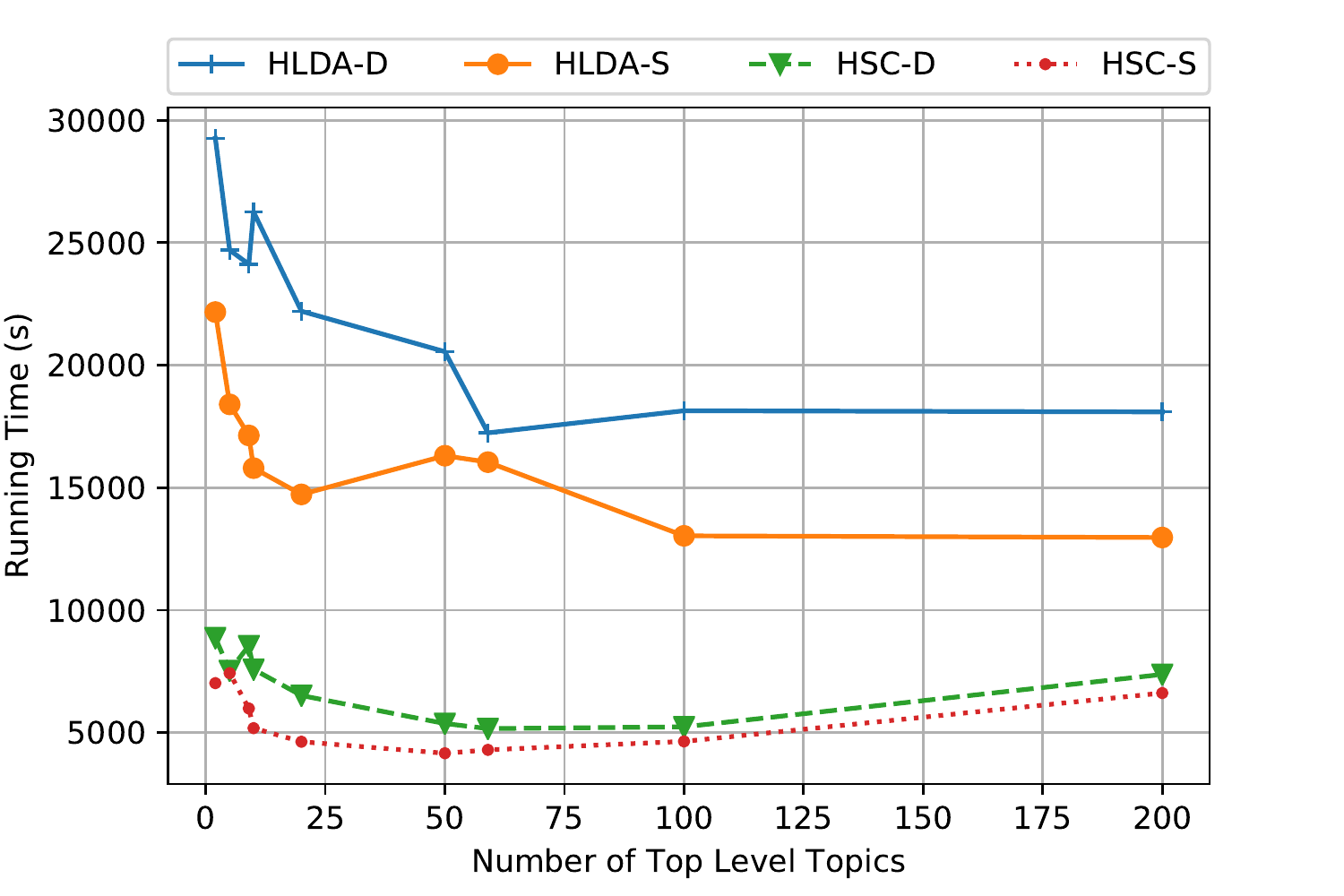}
		\label{fig:marx_time}
	}
	\caption{\small Comparisons of clustering running time}
	\label{fig:hlda_hsc_time}
\end{figure}

\paragraph{\textbf{Comparisons of clustering running time}} Fig. \ref{fig:hlda_hsc_time}
depicts the running time of clustering the BBC News and
Factiva-Marx datasets by different algorithms into two-level clusters 
on a Dell desktop with a quad-core Intel Xeon 2.67 GHz processor and 12 GB RAM.
We choose the top-level clusters numbers $K \in [2, 200]$.

We can see that for both corpora, HSC-S is the fastest, HSC-D is slightly slower,
HLDA-D is the slowest, and HLDA-S is in between HLDA-D and HSC-D. We note
that this result is
expected due to the following two facts: (1) Generating clusters over shorter documents is more time efficient than over longer documents.  (2) Spectral clustering SC is much faster than topic-word distribution clustering LDA.
We also see that when the number of top-level clusters is small, the two-level clustering running time is high.
This is because, for a given corpus, a smaller number of top-level clusters would mean a larger number of second-level clusters, requiring significantly more time
to compute. The turning points are around $K = 20$.
On the other hand, 
when the number
of top-level clusters is larger, the number of second-level clusters is smaller,
which implies a lower time complexity. However, if the number of
sections at the same level is too large, it would make a report harder to read.
Thus, we need to find a balance.

\subsection{Overview reports evaluations}
\label{label:para_setting_lambda}

\paragraph{\textbf{Human judgments}} We recruited human annotators from Amazon Mechanical Turk (AMT) to evaluate six ORPTs generated by NDORGS on, respectively, 0.1, 0.2, and 0.3-summaries of documents in the two corpora, where each report was evaluated by four human annotators. Human judgments are based on the following seven categories: Coherence, UselessText, Redundancy, Referents, OverlyExplicit, Grammatical, and Formatting. 

An ORPT has a high quality if the following seven 
categories all have good ratings: (1) Sentences in the report are coherent. (2) The report does not include useless or confusion text. (3) The report does not contain redundancy information. (4) Common nouns, proper nouns, and pronouns are well referenced in the report. (5) The entity re-mentions are not overly explicit. (6) Grammars are correct. (7) The report is well formatted. 
In particular, we asked AMT annotators to follow the DUC-04 evaluation schema \cite{DUC04questions} in their evaluations. The evaluation scores are provided in Appendix Table \ref{table:human_eval}. 

Fig. \ref{fig:report_AMT} depicts the average scores of human annotators using a 4-point system, with 4 being the best. For the BBC News corpus, Fig. \ref{fig:bbc_AMT} shows that the report generated on 0.3-summaries  outperforms reports generated on 0.2-summaries and 0.1-summaries in all categories except ``OverlyExplicit". Fig. \ref{fig:marx_AMT} shows that the report generated on
0.2-summaries  is better than reports 
generated on 0.3-summaries and 0.1-summaries
in most of the categories; they are``UselessText'', ``Referents'', ``OverlyExplicit'', ``Grammatical'', and ``Formatting''. Note that a larger length ratio of summaries
would help NDORGS generate a better report on BBC News, while a smaller length ratio 
may help NDORGS generate a better report on Factiva-Marx. The reason is likely that the Factiva-Marx corpus contains almost three times more documents than the BBC News corpus, and each document in Factiva-Marx contains on average over ten times larger number of tokens than that in a document from BBC News.
This indicates that for a larger corpus, we may want to use summaries of a smaller length ratio for NDORGS to generate a human preferred overview report.

\begin{figure}[h]
	\centering
	\subfloat[Human Evaluation on BBC News Reports]{
		\includegraphics[width=\columnwidth]{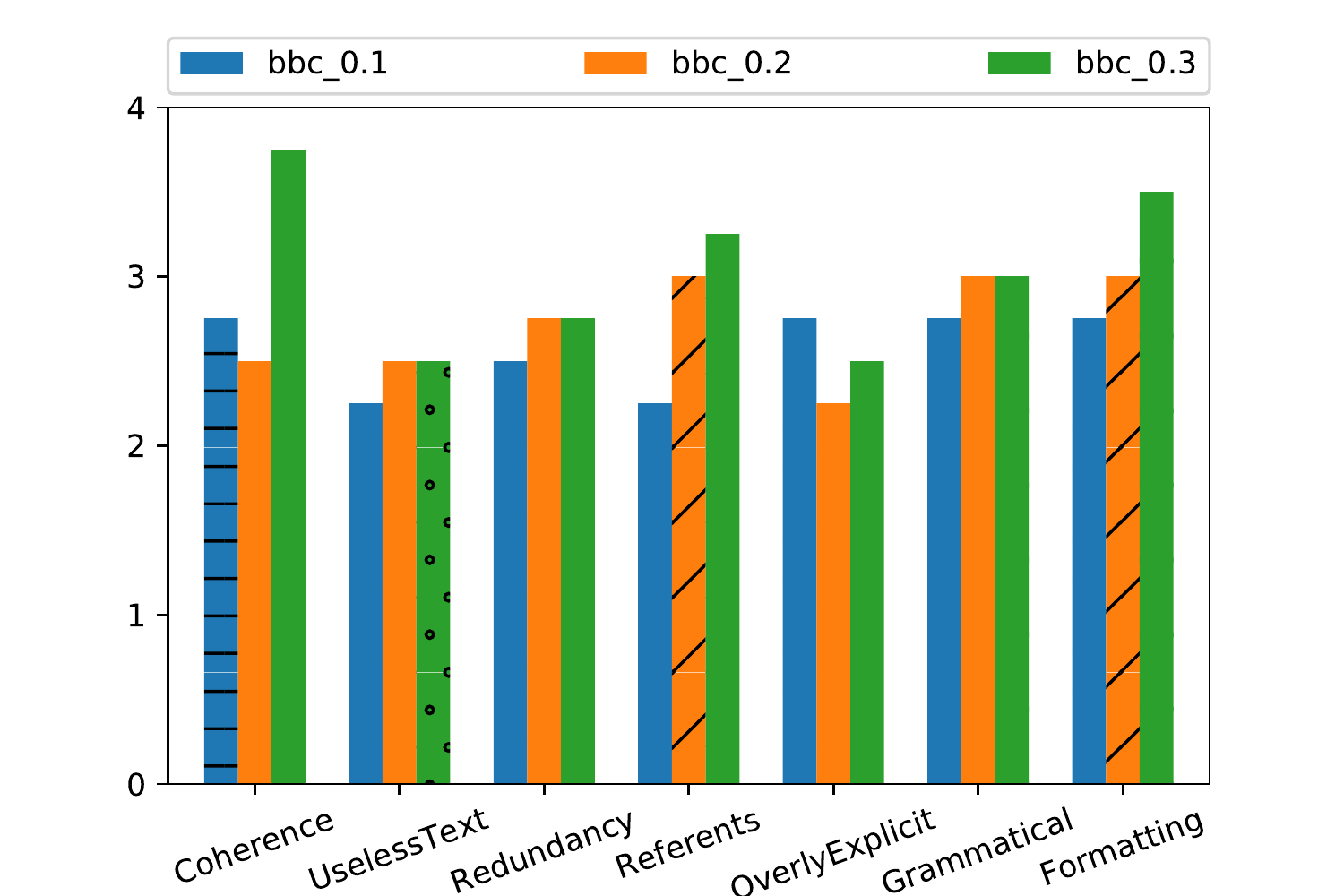}
		\label{fig:bbc_AMT}
	}
	\hfill
	\subfloat[Human Evaluations on Marx Reports]{
		\includegraphics[width=\columnwidth]{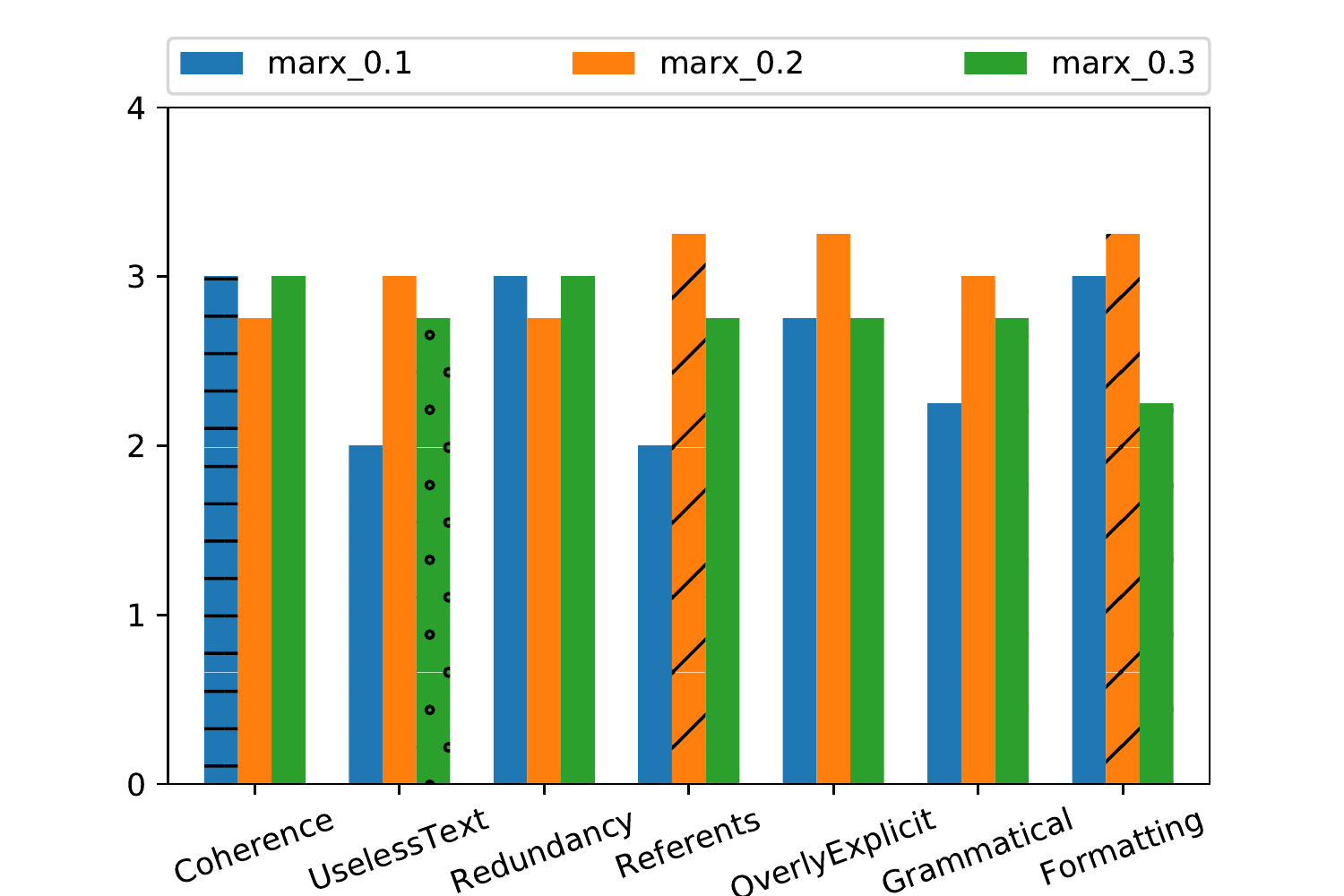}
		\label{fig:marx_AMT}
	}
	\caption{Human Evaluations. \textit{(a) BBC-0.1, BBC-0.2, and BBC-0.3: reports with summary length ratio = 0.1, 0.2, and 0.3 over BBC News. (b) Marx-0.1, Marx-0.2, and Marx-0.3: reports with summary length ratio = 0.1, 0.2, and 0.3 over Factiva-Marx.} }
	\label{fig:report_AMT}
\end{figure}

\begin{figure}[h]
	\centering
	\subfloat[Running time of NDORGS on BBC News]{
		\includegraphics[width=\columnwidth]{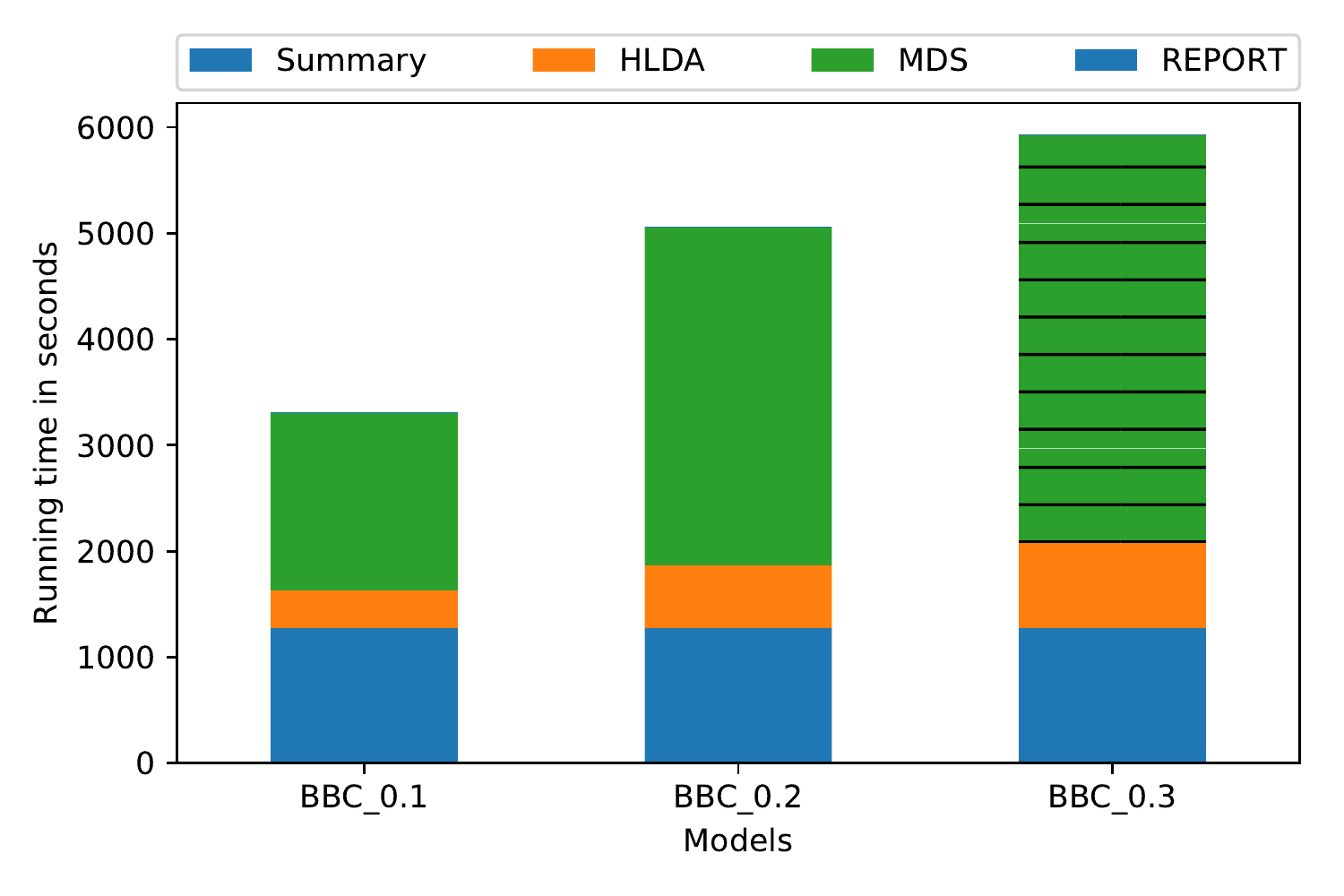}
		\label{fig:bbc_HDORG_time}
	}
	\hfill
	\subfloat[Running time of NDORGS on Factiva-Marx]{
		\includegraphics[width=\columnwidth]{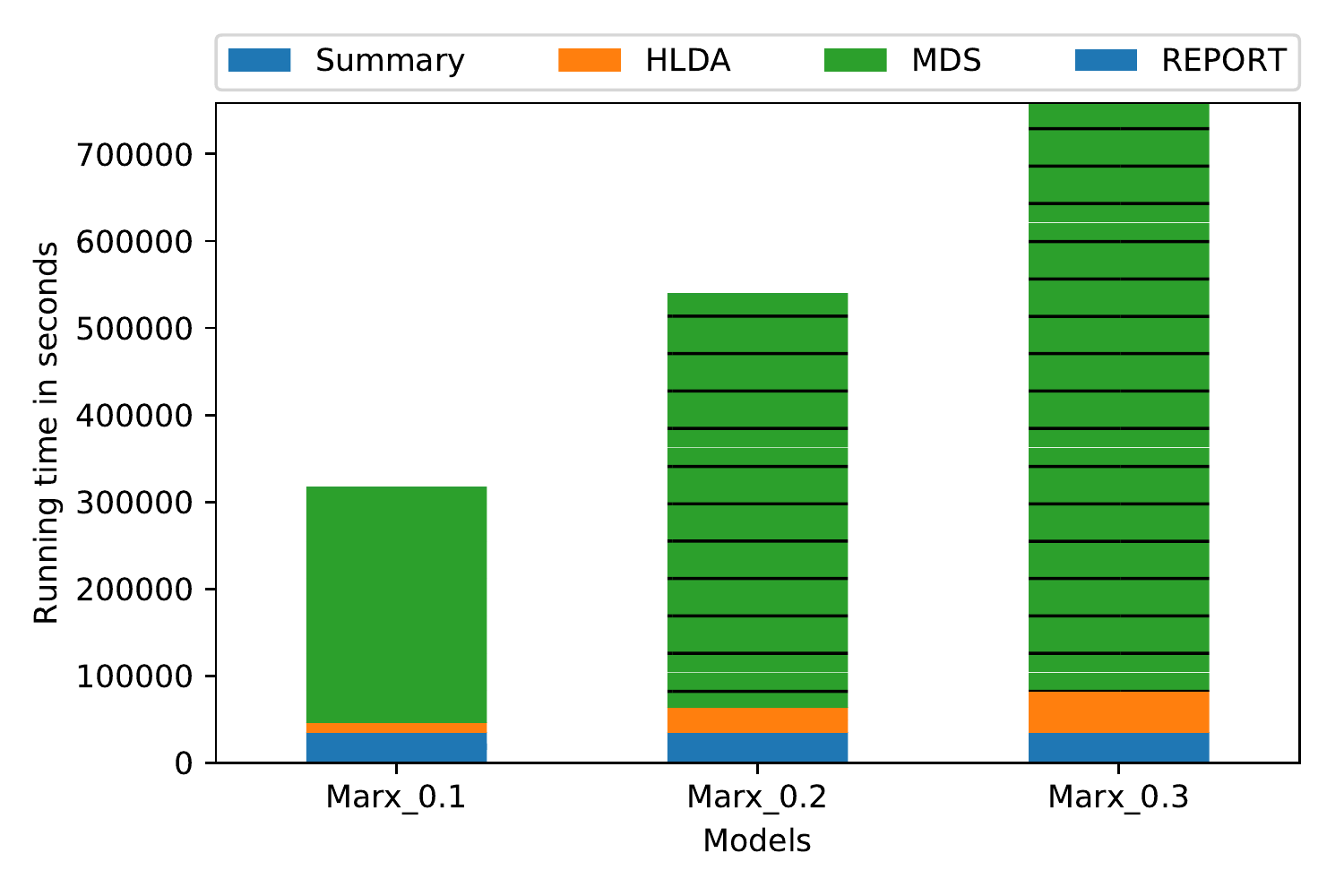}
		\label{fig:marx_HDORG_time}
	}
	\caption{Comparisons of running time}
	\label{fig:HDORG_time}
\end{figure}

\paragraph{\textbf{Time efficiency}}
Fig. \ref{fig:HDORG_time} illustrates the running time of NDORGS on BBC News and Factiva-Marx, with the following results:
\begin{enumerate}
	\item 
	NDORGS incurs, respectively, about 80\% and 56\% more time to generate an overview report on 0.3-summaries and 0.2-summaries than 0.1-summaries (see Fig. \ref{fig:bbc_HDORG_time}).
	\item NDORGS incurs, respectively, over 2 times and 1.8 times
	longer to generate an overview report on 0.3-summaries and 0.2-summaries 
	than it does on 0.1-summaries (see Fig. \ref{fig:marx_HDORG_time}).
	\item NDORGS achieves the best time efficiency on 0.1-summaries
	and the summary length ratios have significantly impacts on time efficiency:
	Working on a larger summary length ratio incurs a longer running time.
\end{enumerate}

The seven evaluation criteria do not include information coverage and
topic diversity, for it is formidable for a human annotator to read through several thousand documents
and summarize information coverage and topic diversity of these documents. 
To overcome this obstable,  we use text mining techniques described in the following two subsections.

\paragraph{\textbf{Information coverage}}
We evaluate information coverage via comparison of the top words in an overview report and the top words in the corresponding corpus (see Section \ref{appendix:topwords} in Appendix). 
Listed below are the summary of the comparison results: 
\begin{enumerate}
	\item For BBC News, over 70\% of the top words in the 
	corpus are also top words in the three overview reports combined, over one-third of the top words in the corpus are top words in the report on 0.1-summaries, over one half of the top words in the corpus are top words in the report on 0.2-summaries as well as in the report on 0.3-summaries, and
	over 80\% of the top 10 words in the corpus
	are top words in each report.
	\item For Factiva-Marx,
	82\% of the top words 
	in the corpus are also top words in the three overview reports combined, 70\% of the top words in the corpus are top words in the report on 0.1-summaries as well as in the report on 0.3-summaries,
	64\% of the top words in the corpus are top words in the report on 0.2-summaries, and the top 12 words in the corpus are top words in each summary. 
	These results indicate that NDORGS is capable of capturing important information of a large corpus.
\end{enumerate}

Let $A$ be a set of top $k$ words from the original corpus and $B$ a set of top $k$ words from a generated report. Let
$S_{k}(A, B) = |A \cap B|/k$
denote the information coverage score of $A$ and $B$.
Then $S_{k}(A, B) \leq 1$ with $S_{k}(A, B) = 1$ being the best possible score.
The information coverage scores for reports over BBC News and Factiva-Marx are listed in Table \ref{table:coverage_scores}. We can see that the report generated on 0.2-summaries achieves the highest information coverage score over BBC News, and the report generated
on 0.1-summaries or 0.3-summaries 
achieves the highest information coverage score over Factiva-Marx.

\begin{table}[h]
	\centering
	\caption{Information-coverage scores}
	\label{table:coverage_scores}
	\begin{tabular}{c | c | c | c}
		\hline
		& $\mathbf{\lambda=0.1}$ & $\mathbf{\lambda=0.2}$ & $\mathbf{\lambda=0.3}$ \\
		\hline
		BBC News & 0.38 & 0.54 & 0.52 \\
		\hline
		Factiva-Marx & 0.70 & 0.64 & 0.70 \\
		\hline
	\end{tabular}	
\end{table}
\vspace*{-3mm}

\paragraph{\textbf{Topic diversity}}
We generate LDA clusters for the original corpus and for the report by treating each sentence as a document in the latter case.
We then evaluate the top words among these clusters using CSD F1-scores to measure topic diversity (see Table \ref{table:diversity_scores}).
We can see that, regarding topic diversity, 
reports generated on 0.2-summaries outperform reports generated on 0.1-summaries and 0.3-summaries for both BBC News and Factiva-Marx.

\begin{table}[h]
	\centering
	\caption{Topic-diversity scores}
	\label{table:diversity_scores}
	\begin{tabular}{c | c | c | c}
		\hline
		& $\mathbf{\lambda=0.1}$ & $\mathbf{\lambda=0.2}$ & $\mathbf{\lambda=0.3}$ \\
		\hline
		BBC News & 0.1278 & 0.1444 & 0.1278 \\
		\hline
		Factiva-Marx & 0.1056 & 0.1167 & 0.1111 \\
		\hline
	\end{tabular}	
\end{table}

\paragraph{\textbf{Overall performance}}
We evaluate the overall performance of overview reports using the following criteria (listed in the order of preference): human evaluation, time efficiency, information coverage, and topic diversity.
We then use Saaty's pairwise comparison 9-point scale and the 
Technique for Order Preference by Similarity to an Ideal Solution (TOPSIS) \cite{hwang1981methods} to determine which length ratio of summaries produces the best overview report. 

Let the three reports
for the same corpus be the three alternatives, denoted by $a_1,a_2,a_3$. Let the human evaluation mean score, running time, information coverage score, and topics diversity score be four criteria, denoted by $c_1,c_2,c_3,c_4$. 
Next, we use Saaty’s pairwise comparison 9-point scale to 
determine weights for each criterion. A weight vector $\bm{w}=\{w_1, w_2, w_3, w_4\}$ is then computed using the Analytic Hierarchy Process (AHP) procedure, where $w_i$ is the weight for criterion $c_i$. A weighted normalization decision matrix $\bm{T}$ is then generated from the normalized matrix $\bm{R}$ and the weight vector $\bm{w}$. The alternatives $a_1$, $a_2$, and $a_3$ are ranked using Euclidean distance and a similarity method (see 
Table \ref{table:hdorg_overall}). 
We can see that the overview report generated on 0.2-summaries achieves the best overall performance on both BBC News and Factiva-Marx. 

\begin{table}[h]
	\centering
	\caption{Overall performance}
	\label{table:hdorg_overall}
	\resizebox{\columnwidth}{!}{%
		\begin{tabular}{ c || c | c | c | c | c }
			\hline
			\textbf{rank} & \textbf{model} & \textbf{human eval.} & \textbf{time} & \textbf{coverage} & \textbf{diversity}  \\
			\hline
			3 & BBC-0.1 & 3.57 & 3310 & 0.38 & 0.1278 \\
			\hline
			\textbf{1} & \textbf{BBC-0.2} & 3.71 & 5060 & 0.54 & 0.1444 \\
			\hline
			2 & BBC-0.3 & 4.03 & 5930 & 0.52 & 0.1278 \\
			\hline
			\hline
			2 & Marx-0.1 & 3.57 & 317023 & 0.70 & 0.1056 \\
			\hline
			\textbf{1} & \textbf{Marx-0.2} & 4.03 & 539474 & 0.64 & 0.1167 \\
			\hline
			3 & Marx-0.3 & 3.75 & 758404 & 0.70 & 0.1111 \\
			\hline
		\end{tabular} 	
	}
\end{table}

\paragraph{\textbf{Sensitivity analysis}}
A stable decision made by TOPSIS is not easily changed when adjusting the weight of the criteria. To evaluate how stable the decision TOPSIS has made, we
carry out sensitivity analyses to measure the sensitivity of weights. For criterion $c_i$, we vary $w_i$ with a small increment $c$ by $w'_i = w_i + c$. We then adjust the weights for other criteria $c_j$ by
\begin{align*}
w'_j = (1-w'_i)w_j/(1-w_i).
\end{align*}

\begin{figure*}[!h]
\centering
	\subfloat[Sensitivity analysis of TOPSIS over BBC News]{
		\includegraphics[width=0.8\textwidth]{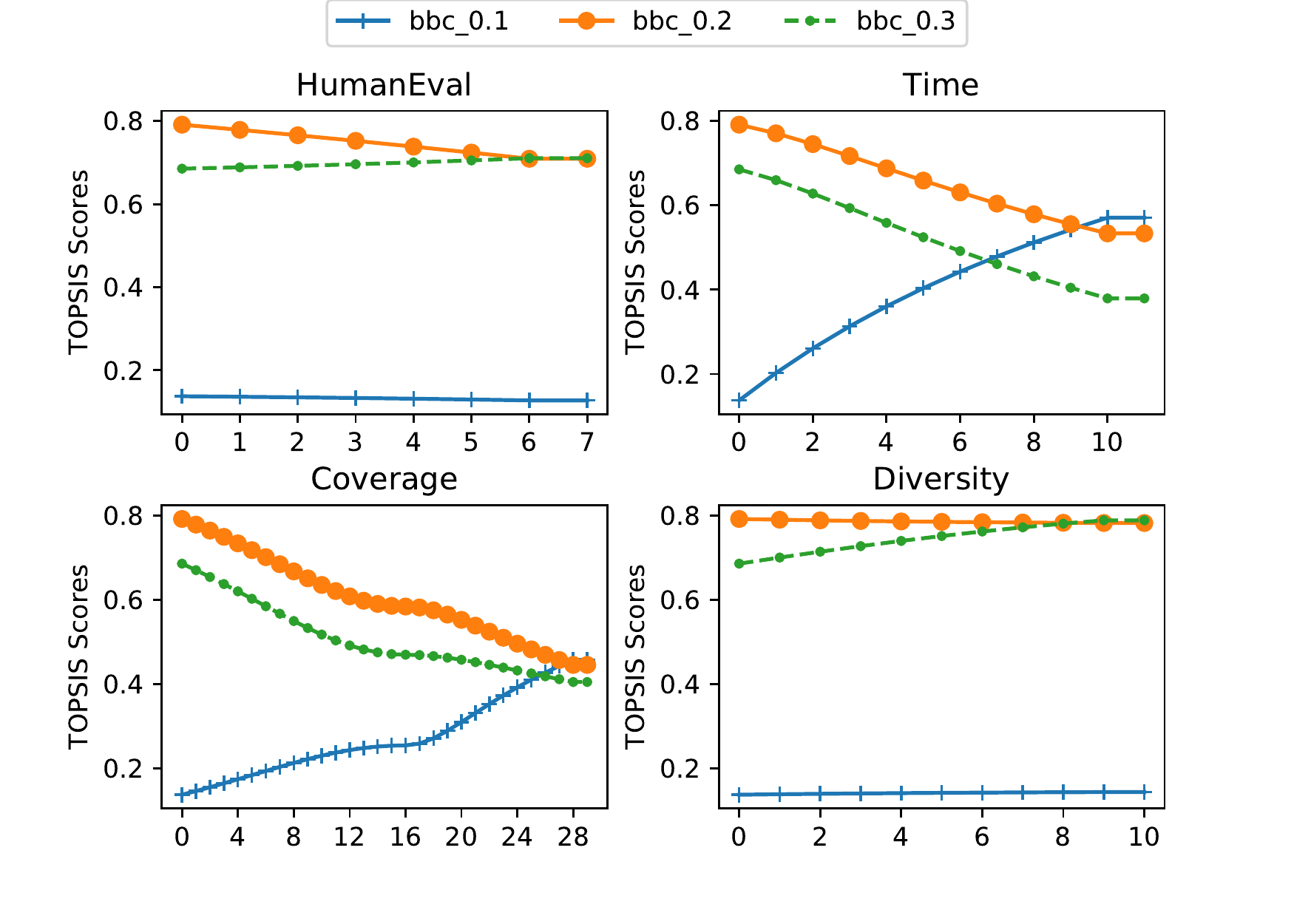}
		\label{fig:bbc_sensi_analy}
	}
	\hfill
	\subfloat[Sensitivity analysis of TOPSIS over  Factiva-Marx]{
		\includegraphics[width=0.8\textwidth]{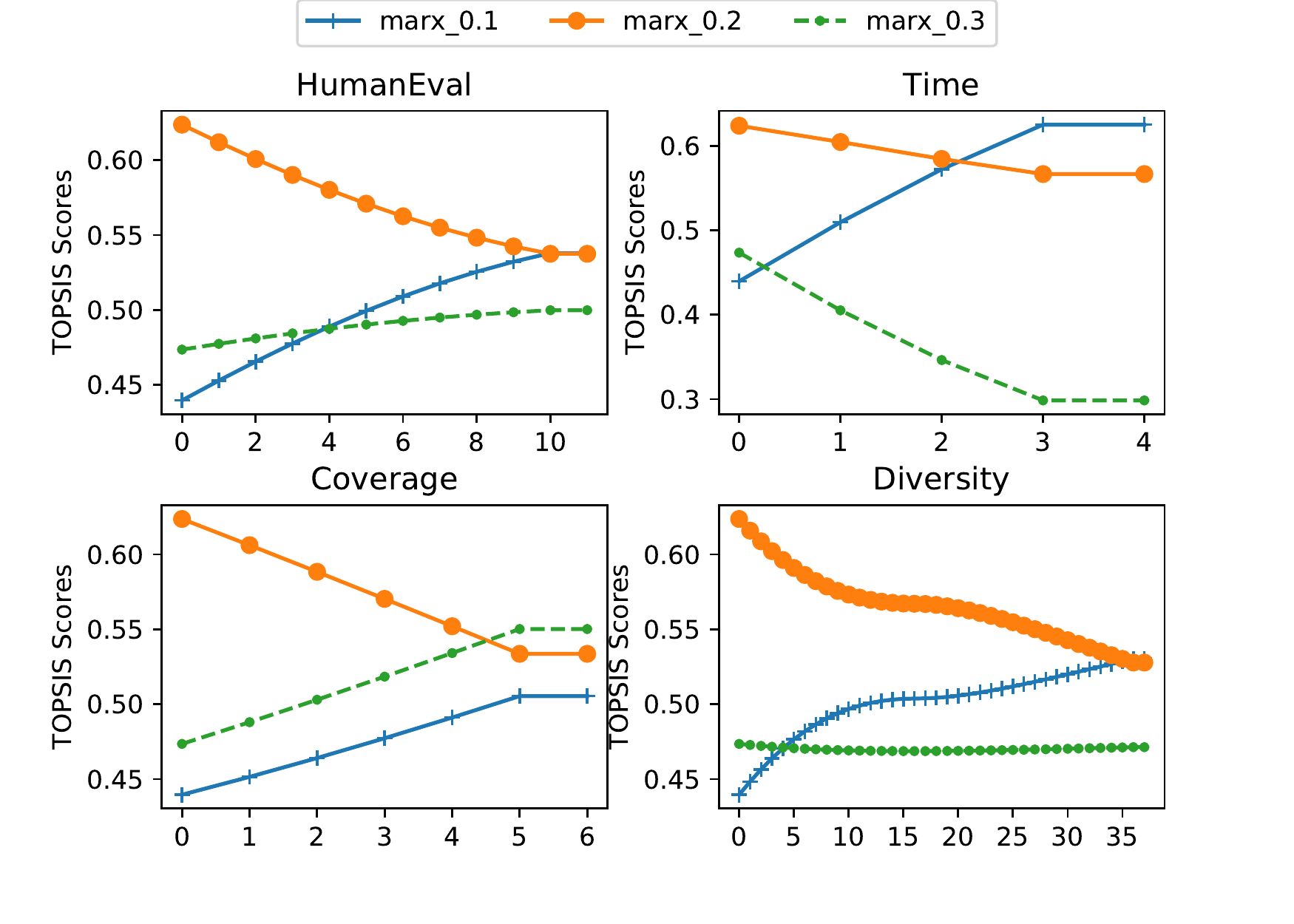}
		\label{fig:marx_sensi_analy}
	}
	\caption{The x-axis indicates the weight of corresponding criterion in increments/decrements of 0.02 each time, and the y-axis shows the new TOPSIS values}
	\label{fig:sensi_analy}
\end{figure*}

We recompute the ranking until another alternative is ranked number one.
Fig. \ref{fig:sensi_analy} depicts the sensitivity analyses results. In both Fig. \ref{fig:bbc_sensi_analy} and Fig. \ref{fig:marx_sensi_analy}, reports generated on 0.2-summaries keep the highest rank while adjusting the weight of criteria of ``Human Evaluation'', ``Time'', ``Coverage'', and ``Diversity'' . Thus, the decision made by TOPSIS is stable over both BBC News and Factiva-Marx.

\section{Trending Graphs}
\label{sec:trending}

\begin{figure*}[!hp]
\centering
\includegraphics[width=0.8\textwidth]{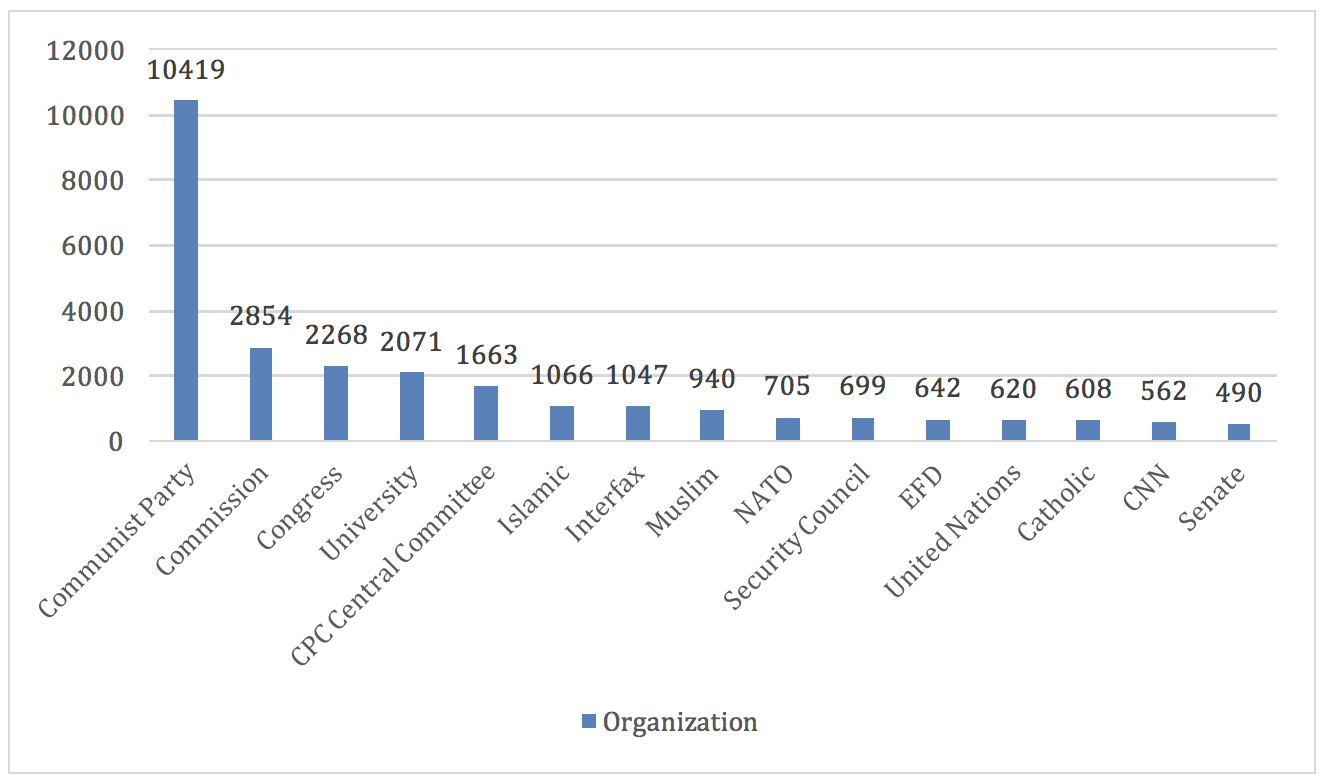}
\caption{The most frequent organizations in Factiva-Marx}
\label{fig:NER1}
\end{figure*}

\begin{figure*}[!hp]
\centering
\includegraphics[width=0.8\textwidth]{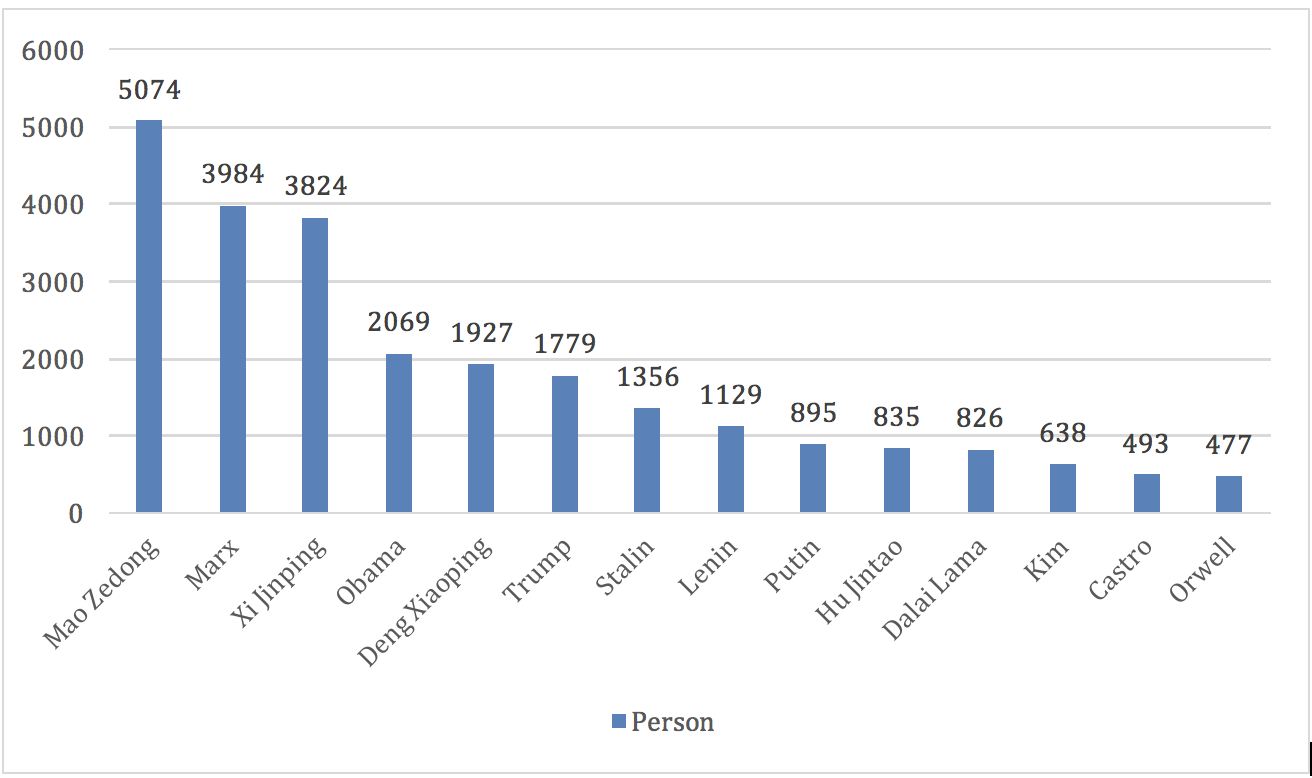}
\caption{The most frequent persons in Factiva-Marx}
\label{fig:NER2}
\end{figure*}

\begin{figure*}[!hp]
\centering
\includegraphics[width=0.8\textwidth]{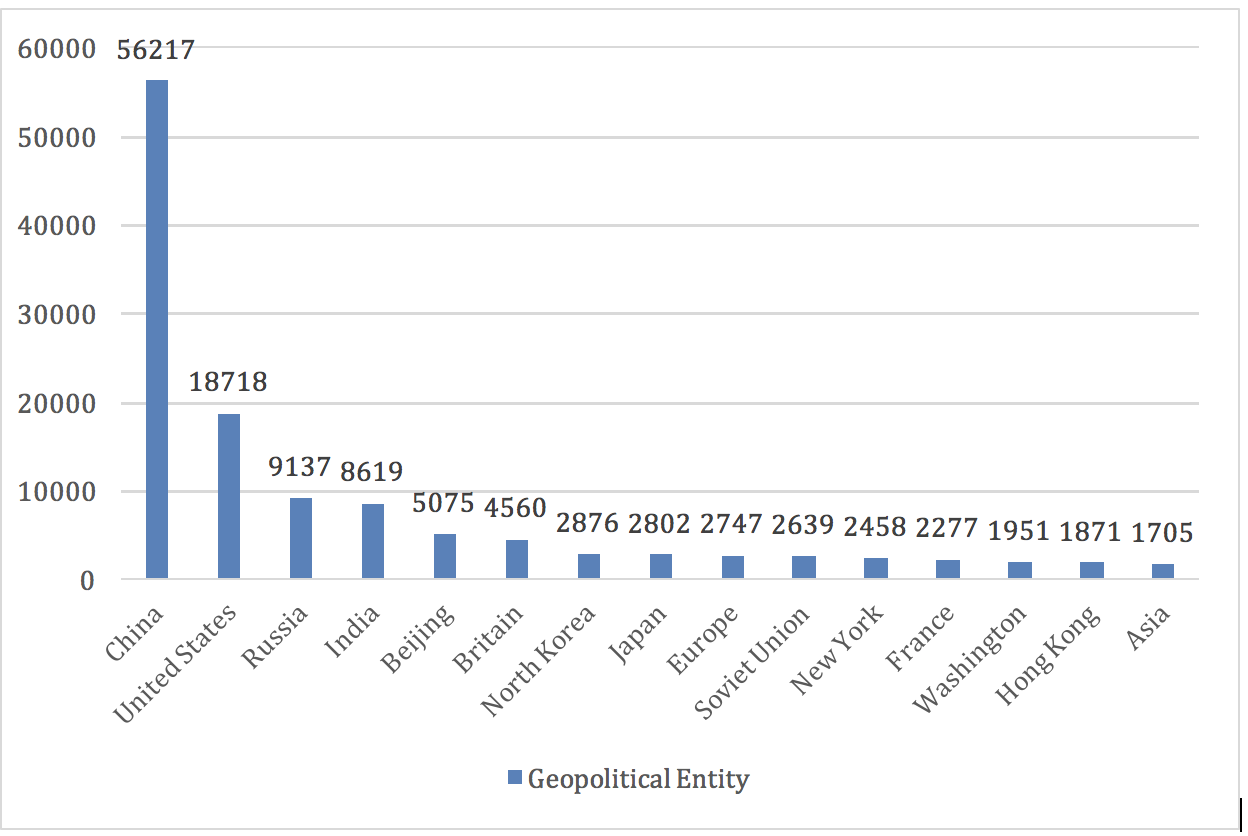}
\caption{The most frequent geopolitical entities in Factiva-Marx}
\label{fig:NER3}
\end{figure*}

In addition to generating the text component of the overview, which is the major part, it is also much desirable to generate trending graphs to provide the reader with an easy visual on
name entities of interests, including organizations, persons, and geopolitical entities. 
In particular, it uses a name-entity-recognition tool (such as nltk.org) to tag name entities
and compute their frequencies. Figs. \ref{fig:NER1}, \ref{fig:NER2}, and \ref{fig:NER3}
are the statistics graphs over the Factiva-Marx dataset.

\begin{figure*}[!hp]
\centering
\includegraphics[width=0.8\textwidth]{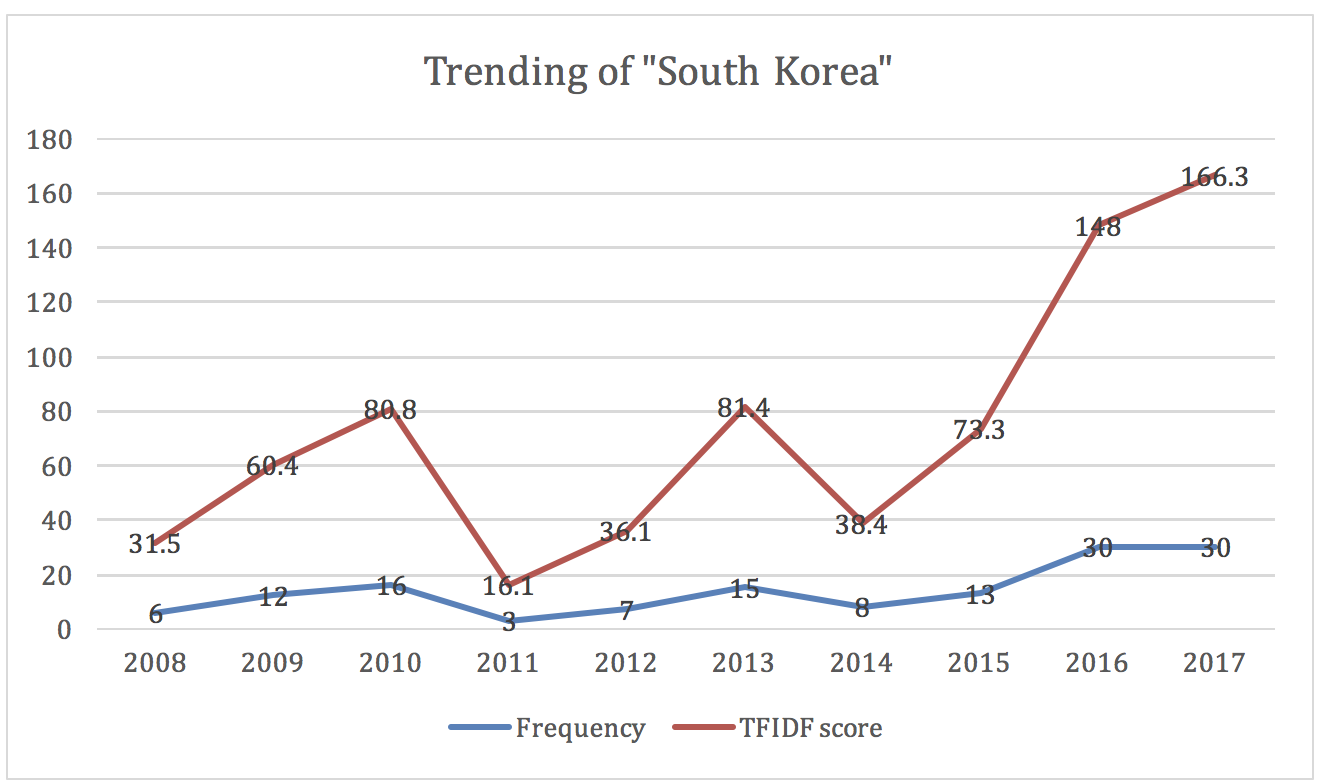}
\caption{The trend of ``South Korea" in Factiva-Marx}
\label{fig:NER4}
\end{figure*}

\begin{figure*}[!hp]
\centering
\includegraphics[width=0.8\textwidth]{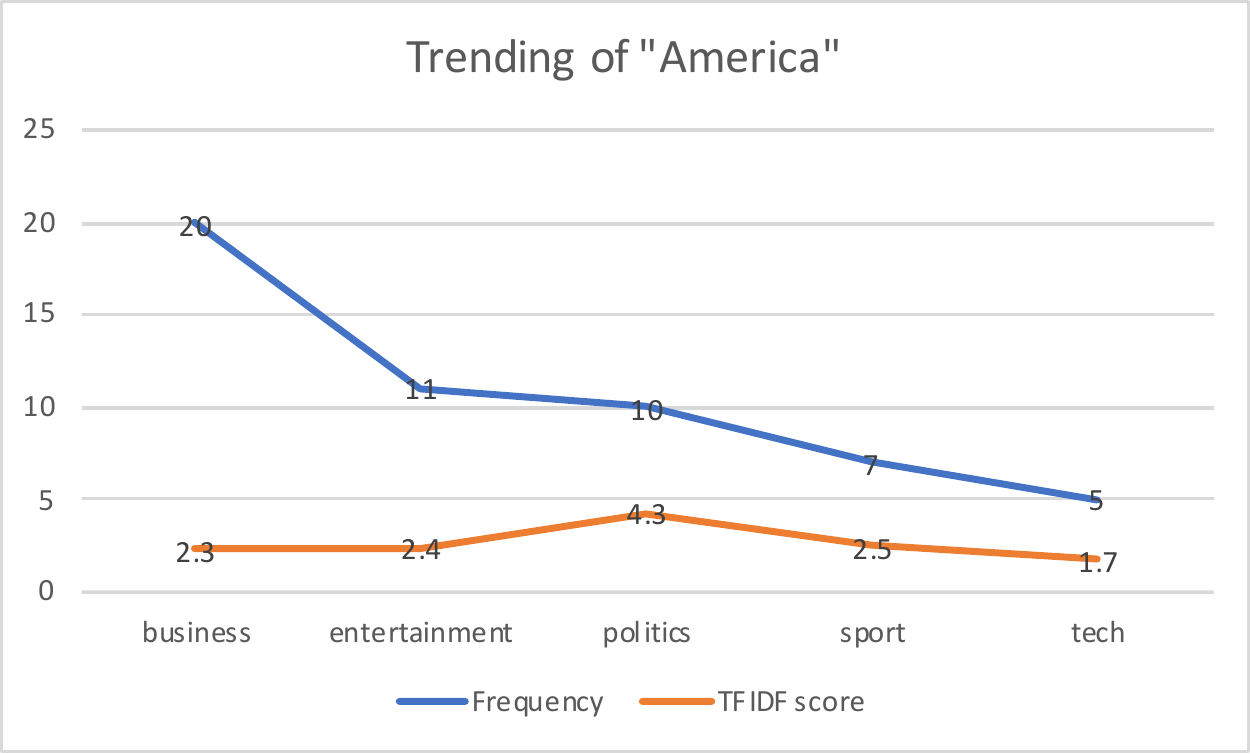}
\caption{The trend of ``America" in BBC News}
\label{fig:bbc_trending_America}
\end{figure*}

For a specific name entity of interests, we also generate a TFIDF score, in addition to its frequency. The TFIDF score of each category is the summation of the TFIDF score of each document with respect to the corpus of articles in that year, which measures its significance. Fig. \ref{fig:NER4} depicts such a trending graph for ``South Korea". It is interesting to note that, while ``South Korea" was both mentioned 30 times in 2016 and 2017, its significance was much higher in 2017 than 2016. Fig. \ref{fig:bbc_trending_America} is the trending graph of
America over BBC News with respect to the six categories of news classifications.
It is interesting to note that, while ``America has one more count
in Entertainment than in Politics, the importance of ``America" in Politics 
is about twice as much as that in Entertainment.

\section{Conclusion}
\label{sec:conclusion}
We presented NDORGS for generating a coherent, well-structured ORPT over a large corpus of documents with acceptable efficiency and accuracy. 
Our experiments
show that the ORPTs generated on 0.2-summaries are the best overall with respect to human evaluations, running time, information coverage, and topic diversity. Sensitivity analysis shows that this result is stable. 

\section*{Acknowledgment}
We thank Dr. Leif Tang and Dr. Ting Wang for making the Factiva-Marx dataset available to us. We are grateful to Dr. Peilong Li for helping some of the experiments.

\appendix
\balance
\section{Appendix}
\label{sec:appendix}
\subsection{Human evaluation scores}

\begin{table}[h]
	\centering
	\caption{Human evaluation Scores, where $\mathbf{C_1, C_2, \ldots, C_7}$ represent,
		respectively, Coherence, UselessText, Redundancy, Referents, OverlyExplicit, Grammatical, and Formatting}
	\label{table:human_eval}
	\begin{adjustbox}{max width=\columnwidth}
		\begin{tabular}{c|c|ccccccc}
			\hline
			\multirow{2}{*}{\textbf{Corpus}} & \multirow{2}{*}{\textbf{Report}} & \multicolumn{7}{c}{\textbf{Human Evaluation Score}} \\
			\cline{3-9}
			& & \textbf{$\mathbf{C_1}$} & \textbf{$\mathbf{C_2}$} & \textbf{$\mathbf{C_3}$} & \textbf{$\mathbf{C_4}$} & \textbf{$\mathbf{C_5}$} & \textbf{$\mathbf{C_6}$} & \textbf{$\mathbf{C_7}$} \\  \hline
			
			\hline
			\hline
			\multirow{12}{*}{BBC News}
			& \multirow{4}{*}{$\lambda=0.1$} 
			&3&1&2&2&2&2&1\\
			&&4&4&4&4&4&4&4\\
			&&3&3&2&1&3&1&2\\
			&&1&1&2&2&2&4&4\\
			
			\cline{2-9}
			
			& \multirow{4}{*}{$\lambda=0.2$}
			&4&3&3&3&0&3&2\\
			&&2&3&3&3&4&4&4\\
			&&1&1&2&3&2&2&4\\
			&&3&3&3&3&3&3&2\\
			
			\cline{2-9}
			
			& \multirow{4}{*}{$\lambda=0.3$}
			&4&3&2&4&4&4&4\\
			&&4&1&3&2&1&1&3\\
			&&3&2&2&3&1&3&3\\
			&&4&4&4&4&4&4&4\\
			
			\hline
			\hline
			
			\multirow{12}{*}{Factiva-Marx}
			& \multirow{4}{*}{$\lambda=0.1$} 
			&3&3&3&1&3&2&3\\
			&&4&1&4&2&4&4&4\\
			&&2&2&3&2&3&1&2\\
			&&3&2&2&3&1&2&3\\
			
			\cline{2-9}
			& \multirow{4}{*}{$\lambda=0.2$}
			&3&2&2&3&3&4&2\\
			&&2&2&2&3&3&3&4\\
			&&2&4&3&3&3&3&3\\
			&&4&4&4&4&4&2&4\\
			
			\cline{2-9}
			
			& \multirow{4}{*}{$\lambda=0.3$}
			&4&3&4&4&3&3&3\\
			&&3&3&3&2&3&3&2\\
			&&2&2&2&2&2&2&2\\
			&&3&3&3&3&3&3&2\\
			\hline
		\end{tabular} 
	\end{adjustbox}
\end{table}

\subsection{Comparisons of top words}
\label{appendix:topwords}

Top words are listed below in the corpus of BBC News and the corpus of Factiva-Marx,
respectively, for comparisons, 
where the first item 
depicts the top 50 words in the original corpus, and the second, third, and fourth items depict, respectively, the top 50 words in the report generated
on 0.1-summaries, 0.2-summaries, and 0.3-summaries, listed in descending order
of keyword scores.
The words in \textbf{bold} are the common top words that occur across all four rows. 
The words with \underline{underlines} are the top words that occur in the first row and two of the other three rows.
The words in \textit{italics} are the top words that occur in the first row and just one of the other three rows.

\paragraph{\textbf{Top word comparisons for BBC News}}
\begin{enumerate}
	\item			\textbf{people}, told, \textbf{best}, \textbf{government}, \textit{time}, \textbf{year}, \textbf{number}, \textbf{three}, \textbf{film},  \textbf{music}, \underline{bbc}, \textit{set}, \textbf{game}, going, \underline{years}, \textbf{labour}, good, well, \textit{top}, \textbf{british}, \textit{european}, \underline{win}, \textit{market}, \textbf{won}, \underline{company}, public, \underline{second}, play, \textit{mobile}, \textit{work}, \textbf{firm}, \underline{blair}, \textbf{games}, minister, \underline{expected}, \underline{england}, \textit{chief}, technology, party, sales, \textit{news}, \textit{plans}, \textit{including}, \textit{help}, \textit{election}, digital, players, director, \textit{economic}, big 
	\item			\textbf{people}, \textbf{best}, \textbf{number}, \textbf{government}, \textbf{film}, \textbf{year}, \textbf{three}, \textbf{game}, howard,  \textbf{music}, london, \textbf{british}, face, biggest, net, action, \textbf{firm}, deal, rise, national, foreign, singer, michael, leader, oil, \underline{blair}, dollar, stock, star, cup, online, future, \textbf{games}, 2004, \textit{work}, \textbf{won}, list, international, coach, \underline{win}, mark, tory, \textbf{labour}, brown, general, prices, \textit{market}, car, \textit{help}, users 
	\item			\textbf{year}, \textbf{people}, \textbf{number}, \textbf{three}, \textbf{best}, \textbf{british}, \textbf{film}, \underline{company}, \textbf{won}, \textbf{labour},  \textbf{music}, net, \underline{bbc}, \textbf{government}, leader, shares, \textit{european}, earlier, chart, third, \textbf{games}, state, \underline{win}, coach, \underline{expected}, \underline{second}, months, political, house, \textit{economic}, \textbf{game}, \underline{years}, team, start, manchester, \underline{england}, \textit{election}, \textit{chief}, international, michael, profit, champion, award, star, announced, service, future, \textbf{firm}, \textit{top}, \textit{news} 
	\item			\textbf{people}, \underline{england}, \textbf{year}, \textbf{film}, \textbf{labour}, boss, \textbf{firm}, despite, \textbf{number}, \textbf{three}, wales, \textbf{british}, nations, \textbf{best}, \underline{company},  \textbf{music}, \underline{blair}, \textit{set}, record, oil, \textit{time}, \underline{years}, \textbf{won}, prices, \textit{plans}, net, online, \textit{including}, films, \underline{bbc}, court, \textbf{games}, \textbf{game}, brown, david, \textbf{government}, \underline{expected}, club, action, beat, total, group, unit, firms, rules, \textit{mobile}, \underline{second}, analysts, future, computer 
\end{enumerate}

\paragraph{\textbf{Top word comparisons for Factiva-Marx}}
\begin{enumerate}
	\item			\textbf{party}, \textbf{chinese}, \textbf{china}, \textbf{political}, \textbf{people}, \textbf{communist}, \textbf{economic}, \textbf{national}, \textbf{state}, \textbf{government}, \textbf{years}, \textbf{social}, \textit{great}, \underline{time}, \textbf{rights}, \underline{development}, \underline{international}, \textbf{president}, \textbf{central}, \textbf{war}, \textbf{north}, \underline{university}, \textbf{power}, \textbf{united}, work, \underline{country}, \textbf{foreign}, \textit{global}, military, \underline{history}, \underline{south}, \textbf{marxism}, \textbf{human}, \textbf{western}, \textit{soviet}, well, system, \textbf{mao}, \textit{american}, \textbf{news}, \underline{public}, \textit{cultural}, long, \underline{states}, \textit{countries}, three, left, \textbf{media}, british, including 
	\item			\textbf{party}, \textbf{china}, \textbf{chinese}, \textbf{communist}, \textbf{political}, \textbf{years}, \textbf{economic}, \textbf{rights}, \textbf{human}, \textbf{president}, \textbf{people}, \textbf{national}, year, \textbf{state}, leaders, \textbf{government}, \textbf{central}, \textit{countries}, \textbf{news}, \textbf{social}, \underline{country}, leader, \underline{time}, \textbf{foreign}, \textbf{power}, \textbf{north}, nuclear, top, \textbf{marxism}, ideological, led, \textbf{media}, \textbf{war}, beijing, \textbf{western}, \textbf{united}, \underline{development}, \textit{soviet}, \textbf{mao}, \underline{states}, \underline{history}, \underline{university}, capitalism, official, market, officials, march, korea, democracy, \underline{south} 
	\item			\textbf{china}, \textbf{party}, \textbf{communist}, \textbf{chinese}, \textbf{political}, \textbf{economic}, \textbf{years}, \textbf{rights}, \textbf{president}, \textbf{people}, \textbf{central}, \textbf{state}, \textbf{social}, \textbf{united}, \textbf{north}, beijing, \textbf{western}, \textbf{news}, \textbf{media}, \textbf{mao}, cpc, \textbf{war}, \textbf{human}, anniversary, \underline{public}, members, \underline{country}, jinping, leader, \underline{states}, \textbf{government}, \underline{south}, \textbf{marxism}, democratic, \textbf{national}, \textbf{power}, year, \textbf{foreign}, \textit{american}, education, \underline{international}, july, nuclear, day, book, leadership, committee, leaders, copyright, study 
	\item			\textbf{china}, \textbf{party}, \textbf{communist}, \textbf{chinese}, \textbf{economic}, \textbf{years}, \textbf{people}, \textbf{political}, \textbf{human}, \textbf{news}, \textbf{state}, \textbf{social}, \textbf{government}, \textbf{central}, \textbf{national}, leader, \textbf{president}, \textbf{media}, \textit{cultural}, \textbf{rights}, \textbf{mao}, \textbf{power}, \underline{development}, year, \underline{international}, \underline{university}, leaders, \underline{history}, \textbf{united}, beijing, copyright, socialist, \textit{global}, \textit{great}, top, nation, universities, \textbf{western}, revolution, nuclear, \textbf{foreign}, \underline{public}, agency, \textbf{marxism}, \underline{time}, members, congress, \textbf{war}, change, \textbf{north} 
\end{enumerate}

\end{document}